\documentclass[sigconf]{acmart}
\renewcommand\footnotetextcopyrightpermission[1]{} % removes footnote with conference information in first column
\def\BibTeX{{\rm B\kern-.05em{\sc i\kern-.025em b}\kern-.08emT\kern-.1667em\lower.7ex\hbox{E}\kern-.125emX}}
\usepackage{setspace}
\usepackage{wrapfig}
\usepackage{graphicx}
\usepackage{times}
\usepackage{epsfig}
\usepackage{graphicx}
\usepackage{amsmath}
  
\usepackage{amssymb}
\usepackage[ruled]{algorithm2e}
\usepackage{multirow}
\usepackage{indentfirst}
\usepackage{graphicx}
\usepackage{float}
\usepackage{dirtree}
\usepackage{amsmath,amssymb}
\usepackage{setspace}
\usepackage{mathtools}
\usepackage{float}
\usepackage[normalem]{ulem}
\usepackage{array}
\usepackage{epstopdf}
\usepackage{amsmath}
\usepackage{booktabs}
\usepackage{url}
\usepackage{float}

\begin{document}

\title{MangaGAN: Unpaired Photo-to-Manga Translation Based on\\The Methodology of Manga Drawing }

\author{Hao Su$^1$, Jianwei Niu$^{1,2,3*}$, Xuefeng Liu$^1$, Qingfeng Li$^1$, Jiahe Cui$^1$, and Ji Wan$^1$}
\affiliation{
$^1$State Key Lab of VR Technology and System, School of Computer Science and Engineering, Beihang University \\ 
$^2$Industrial Technology Research Institute, School of Information Engineering, Zhengzhou University\\
$^3$Hangzhou Innovation Institute, Beihang University\\
\{bhsuhao, niujianwei, liu\_xuefeng, liqingfeng, cuijiahe, wanji\}@buaa.edu.cn\\
$ \ $ \\
\textbf{\LARGE{This paper has been accepted by AAAI 2021.}}}
\thanks{$^*$The corresponding author. }

\renewcommand{\shortauthors}{}

\begin{abstract}
Manga is a world popular comic form originated in Japan, which typically employs black-and-white stroke lines and geometric exaggeration to describe humans' appearances, poses, and actions. In this paper, we propose MangaGAN, the first method based on Generative Adversarial Network (GAN) for unpaired photo-to-manga translation. Inspired by how experienced manga artists draw manga, MangaGAN generates the geometric features of manga face by a designed GAN model and delicately translates each facial region into the manga domain by a tailored multi-GANs architecture. For training MangaGAN, we construct a new dataset collected from a popular manga work, containing manga facial features, landmarks, bodies and so on. Moreover, to produce high-quality manga faces, we further propose a structural smoothing loss to smooth stroke-lines and avoid noisy pixels, and a similarity preserving module to improve the similarity between domains of photo and manga. Extensive experiments show that MangaGAN can produce high-quality manga faces which preserve both the facial similarity and a popular manga style, and outperforms other related state-of-the-art methods.
\end{abstract}

%\keywords{Manga, Image translation, Generative adversarial network}

\begin{teaserfigure}
\vspace{-0.3cm}
\centering
  \includegraphics[width=0.9\textwidth]{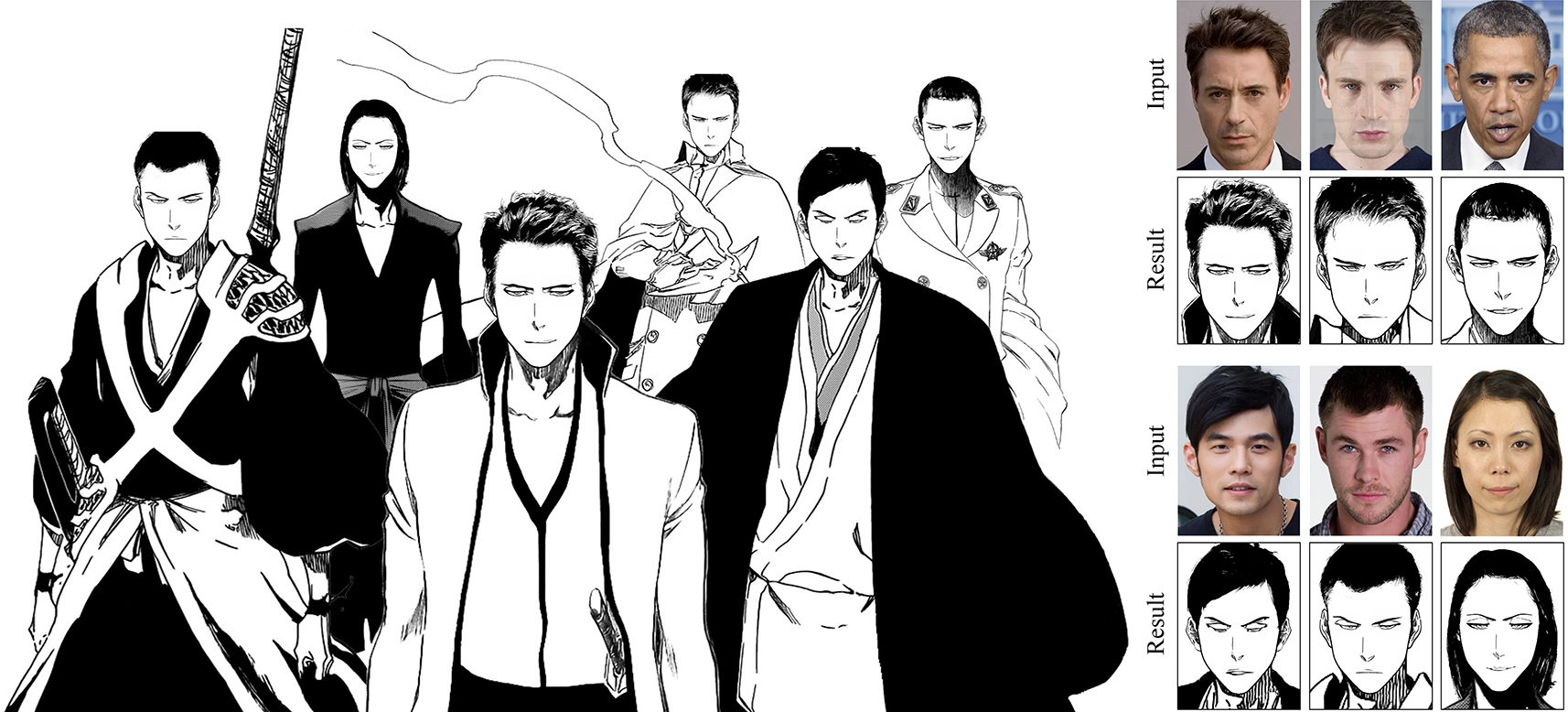}
   \vspace{-0.2cm}
  \caption{\emph{Left:} the combination of manga faces we generated and body components we collected from a popular manga work \emph{Bleach} \cite{Bleach}, which shows a unified style with a strong attractiom.  \emph{Right:} the input frontal face photos and our results, where our method can effectively endow output results with both the facial similarity and the target manga style.}
  \label{fig:teaser}
\vspace{0.2cm}
\end{teaserfigure}
\maketitle

\section{Introduction}
\label{sec:introduction}

\emph{Manga}, originated in Japan, is a worldwide popular comic form of drawing on serialized pages to present long stories. Typical manga is printed in black-and-white (as shown in Fig.~\ref{fig:teaser} left), which employs abstract stroke lines and geometric exaggeration to describe humans' appearances, poses, and actions. Professional manga artists usually build up personalized drawing styles during their careers, and their styles are hard to be imitated by other peers. Meanwhile, drawing manga is a time-consuming process, and even a professional manga artist requires several hours to finish one page of high-quality work.

As an efficient approach to assist with manga drawing, automatically translating a face photo to manga with an attractive style is much desired. This task can be described as the image translation that is a hot topic in the computer vision field. In recent years, deep learning based image translation has made significant progress and derived a series of systematic methods. Among the examples are the \emph{Neural Style Transfer} (NST) methods (e.g.,\cite{Gatys,Markov,stylebank,Lff}) which use tailored CNNs and objective functions to stylize images, the \emph{Generative Adversarial Network} (GAN)\cite{gan} based methods (e.g., \cite{pixel2pixel, cyclegan,UINT}) which work well for mapping paired or unpaired images from the original domain to the stylized domain.

Although these excellent works have achieved good performances in their applications, they have difficulties to generate a high-quality manga due to the following four \emph{challenges}. First, in the manga domain, humans' faces are abstract, colorless, geometrically exaggerated, and far from that in the photo domain. The facial correspondences between the two domains are hard to be matched by networks. Second, the style of manga is more represented by the structure of stroke lines, face shape,  and facial features' sizes and locations. Meanwhile, for different facial features, manga artists always use different drawing styles and locate them with another personalized skill. These independent features (i.e., appearance, location, size, style) are almost unable to be extracted and concluded by a network simultaneously. Third, a generated manga has to faithfully resemble the input photo to keep the identity of a user without comprising the abstract manga style. It is a challenge to keep both of them with high performances. Forth, the training data of manga is difficult to collect. Manga artists often use local storyboards to show stories, which makes it difficult to find clear and complete manga faces with factors such as covered by hair or shadow, segmented by storyboards, low-resolution and so on. Therefore, related state-of-the-art methods of image stylization (e.g., \cite{Gatys,Markov,pixel2pixel,cyclegan,dia,headshot,apdrawinggan}) are not able to produce desired results of manga\footnote{Comparison results as shown in Figure \ref{fig:sotaNST} and \ref{fig:sotaGAN} of experiments.}.

To address these challenges, we present MangaGAN, the first GAN-based method for translating frontal face photos to the manga domain with preserving the attractive style of a popular manga work \emph{Bleach} \cite{Bleach}. We observed that an experienced manga artist generally takes the following steps when drawing manga: first outlining the exaggerated face and locating the geometric distributions of facial features, and then fine-drawing each of them. MangaGAN follows the above process and employs a multi-GANs architecture to translate different facial features, and to map their geometric features by another designed GAN model. Moreover, to obtain high-quality results in an unsupervised manner, we present a Similarity Preserving (SP) module to improve the similarity between domains of photo and manga, and leverage a structural smoothing loss to avoid artifacts.

To summarize, our main contributions are three-fold:
\begin{itemize}\setlength{\itemsep}{0pt}

\item We propose MangaGAN, the first GAN-based method for unpaired photo-to-manga translation. It can produce attractive manga faces with preserving both the facial similarity and a popular manga style. MangaGAN uses a novel network architecture by simulating the drawing process of manga artists, which generates the exaggerated geometric features of faces by a designed GAN model, and delicately translates each facial region by a tailored multi-GANs architecture.
\item We propose a similarity preserving module that effectively improves the performances on preserving both the facial similarity and manga style. We also propose a structural smoothing loss to encourage producing results with smooth stroke-lines and less messy pixels.
\item We construct a new dataset called MangaGAN-BL (containing manga facial features, landmarks, bodies, etc.), collected from a world popular manga work {\emph{Bleach}}. Each sample has been manually processed by cropping, angle-correction, and repairing of disturbing elements (e.g, hair covering, shadows). MangaGAN-BL will be released for academic use.
\end{itemize}

\section{Related Work}
Recent literature suggests two main directions with the ability to generate manga-like results: neural style transfer, and GAN-based cross-domain translation.

\subsection{Neural style transfer}
The goal of neural style transfer (NST) is to transfer the style from an art image to another content target image. Inspired by the progress of CNN, Gatys et al. \cite{Gatys} propose the pioneering NST work by utilizing CNN's power of extracting abstract features, and the style capture ability of Gram matrices \cite{gram}. Then, Li and Wand \cite{Markov} use the Markov Random Field (MRF) to encode styles, and present an MRF-based method (CNNMRF) for image stylization.
Afterward, various follow-up works have been presented to improve their performances on visual quality \cite{dia, HeadStyleTransfer,SeparatingNST_CVPR2018,Arbitrary_NST_CVPR2018,Stroke_NST_ECCV2018, text_CVPR2018}, generating speed \cite{stylebank,Lff,Arbitraryfast_CVPR2017,NSTCVPR2018,universalNST_NIPS2017}, and multimedia extension \cite{Coherent, NST_VIDEO_CVPR2017,StylizeQR,artcoder,stereoscopicNST_CVPR2018}.

Although these methods work well on translating images into some typical artistic styles, e.g., oil painting, watercolor, they are not good at producing black-and-white manga with exaggerated geometry and discrete stroke lines, since they tend to translate textures and colors features of a target style and preserve the structure of the content image.

\begin{figure*}[ht]
%\vspace{-0.5cm}
\centering
\includegraphics[width=6.4 in]{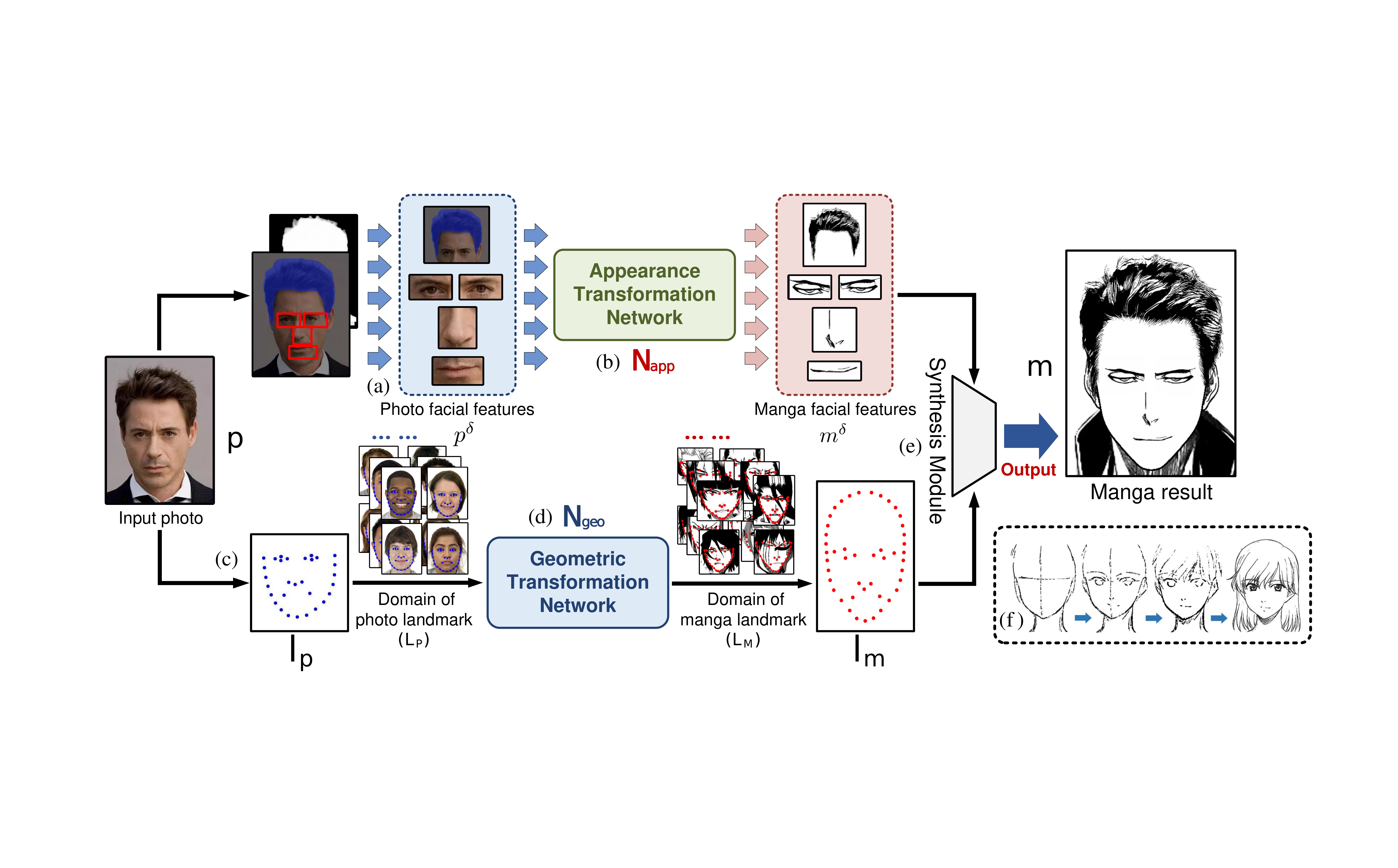}

\caption{Overall pipeline of MangaGAN. Inspired by the prior knowledge of manga drawing, MangaGAN consists of two branches: one branch learns the geometric mapping by a Geometric Transformation Network (GTN); the other branch learns the appearance mapping by an Appearance Transformation Network (ATN). On the end, a Synthesis Module is designed to fuse them and end up with the manga face.}
\label{fig:overview}

\end{figure*}
\subsection{GAN-based cross-domain translation} Many GAN-based cross-domain translation methods work well on image stylization, whose goal is to learn a mapping from a source domain to a stylized domain. There are a series of works based on GAN \cite{gan} presented and applied for image stylization. Pix2Pix \cite{pixel2pixel} first presents a unified framework for image-to-image translation based on conditional GANs \cite{conditionalgan}. BicycleGAN \cite{BicycleGAN} extends it to multi-modal translation. Some methods including CycleGAN \cite{cyclegan}, DualGAN \cite{Dualgan}, DiscoGAN \cite{discoGAN}, UNIT \cite{UINT}, DTN \cite{DTN} etc. are presented for unpaired one-to-one translation. MNUIT \cite{Munit}, startGAN \cite{stargan} etc. are presented for unpaired many-to-many translation.
 %Among them, the most related to our work is CycleGAN\cite{cyclegan}, a typical model of unpaired image translation. It introduces a cycle-consistency loss to improve the performance in mapping two domains, which has been widely used in the tasks of unpaired image translation.

The methods mentioned above succeed in translation tasks that are mainly characterized by color or texture changes only (e.g., summer to winter, and apples to oranges). For photo-to-manga translation, they fail to capture the correspondences between two domains due to the abstract structure, colorless appearance, and geometric deformation of manga drawing.

Besides the above two main directions, there are also some works specially designed for creating artistic facial images. They employ techniques of Non-photorealistic rendering (NPR), data-driven synthesizing, computer graphics, etc., and have achieved much progress in many typical art forms, e.g., caricature and cartoon \cite{carigans,warpgan, cartoon2photo,cartoonTIP2016,carigan2018,cartoongan,1990SIGGRAPH,2006real}, portrait and sketching \cite{apdrawinggan,portraitbenchmarking,potraitTOG2016,renderingportraits, stylitTOG2016, Portraitvideo,sketchTOG2013,sketch2018,sketchNNLS2015,sketchTIP2015,sketchTIP2018}.
However, none of them involve the generation of manga face.
% due to the four challenges aforementioned in Section \ref{sec:introduction}.

\section{Method}
\subsection{Overview}
  Let $P$ and $M$ indicate the face photo domain and the manga domain respectively, where no pairing exists between them. Given an input photo $p \in P$, our MangaGAN learns a mapping $\Psi  $ : $P \rightarrow M$ that can transfer $p$ to a sample $m = \Psi (p)$, $m \in M$, while endowing $m$ with manga style and facial similarity.

 As shown in Figure~\ref{fig:overview}(f), our method is inspired by the prior knowledge that how experienced manga artists doing drawing manga: first outline the exaggerated face and locate the geometric distributions of facial features, and finally do the fine-drawing. Accordingly, MangaGAN consists of two branches: one branch learns a geometric mapping $\Psi_{geo}$ by a Geometric Transformation Network (GTN) $N_{{geo}}$ which adopted to translate the facial geometry from $P$ to $M$ [Figure \ref{fig:overview}(d)]; the other branch learns an appearance mapping $\Psi_{app}$ by an Appearance Transformation Network (ATN) $N_{app}$ [Figure \ref{fig:overview}(b)] which used to produce components of all facial features. At the end, a Synthesis Module is designed to fuse facial geometry and all components, and end up with the output manga $m \in M$ [Figure \ref{fig:overview}(e)]. Then, we will detail the ATN, the GTN, and the Synthesis Module in Section \ref{section:ATN}, Section \ref{section:GTN}, and Section \ref{section:fusion} respectively.

 %Meriting of previous works \cite{carigans, transgaga} that separate the geometry and appearance features, and \cite{apdrawinggan,tpgan,cartoonTIP2016} that introduce local networks for image translation, our MangaGAN employs a tailored GTN to get mange's geometric exaggeration, and design an ATN with multi-GAN architecture, instead of a global one, to better capture local features. Furthermore, we present an Similarity Preserving (SP) module and design effective training strategies to optimize the performance of manga generation.
%based on the methodology of manga drawing [Figure \ref{fig:overview}(d)]

 %Then we will detail the PCA module on the first branch, the multi-GAN architecture on the other branch, and the fusion network, in Section \ref{section:PCA}, \ref{section:GAN}, and \ref{section:fusion}, respectively.

\subsection{Appearance transformation network}
\label{section:ATN}

\begin{figure}[t]
\centering
\includegraphics[width=3.2 in]{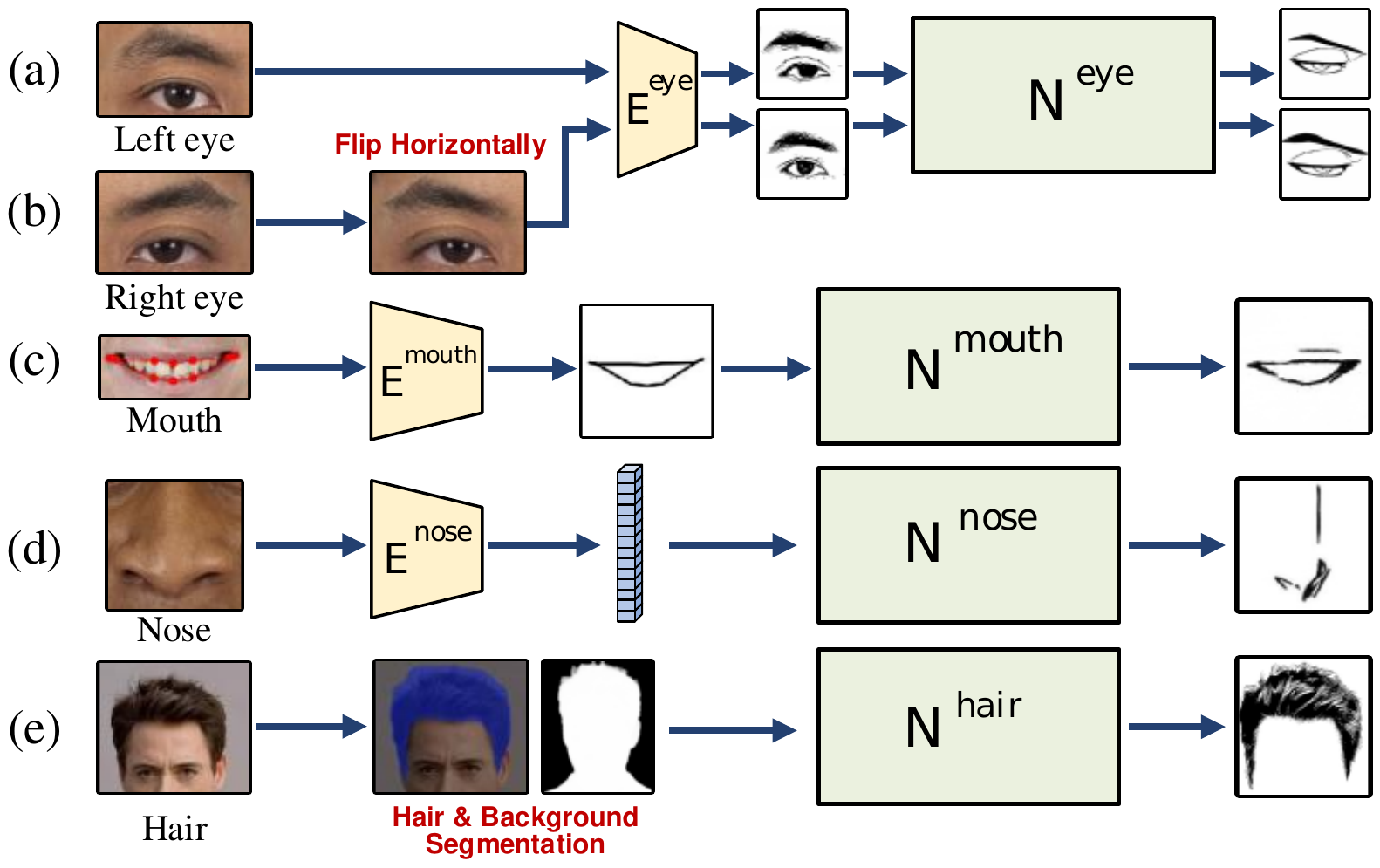}

\caption{ATN is a network with multi-GANs architecture, consists of four local GANs, designed to translate each facial region respectively. Moreover, we tailor different training strategies and encoders to improve their performances. }
\label{fig:multigan}

\end{figure}

As shown in Figure \ref{fig:multigan}, ATN $N_{app}$ is a network with multi-GAN architecture includes a set of four locals GANs, $N_{app}$$=$$\{N^{eye}$, $N^{nose}$, $N^{mouth}$, $N^{hair}\}$, where $N^{eye}$, $N^{nose}$, $N^{mouth}$, and $N^{hair}$ are respectively trained for translating facial regions of eye, nose, mouth, and hair, from the input $p\in P$ to the output $m \in M$.

%\textbf{Architecture of local GANs}. Let $p^{\delta}$ and $m^{\delta}$ express one of target facial regions of $p$ and $m$ respectively, $\delta \!\! \in$$\{$eye, nose, mouth, hair$\}$. The goal of a local GAN $N^{\delta}$ is to learn a mapping $m^{\delta}$$=$$\Psi_{app}^{\delta}(p^{\delta})$.
\vspace{-0.2cm}

\subsubsection{$\!\!\!$Translating regions of eyes and mouths}

\label{eyes_and_mouth}

%\textbf{Baseline architecture of $\mathbf{N^{eye}}$ and $\mathbf{N^{mouth}}$.}
 Eyes and mouths are the critical components of manga faces but are the hardest parts to translate, since they are most noticed, error sensitive, and vary with different facial expressions. For ${N^{eye}}$ and ${N^{mouth}}$, for better mapping the unpaired data, we couple it with a reverse mapping, inspired by the network architecture of CycleGAN \cite{cyclegan}. Accordingly, the baseline architecture of $N^{\delta}$$(\delta$$\in$$\{$eye, mouth$\})$ includes the forward$\,$/$\,$backward generator $G^{\delta}_{M}\,$/$\,G^{\delta}_{P}$ and the corresponding discriminator $D^{\delta}_{P}\,$/$\,D^{\delta}_{M}$. $G^{\delta}_{M}$ learns the mapping $\Psi^\delta_{app} \!\! : p^{\delta} \!\!\! \rightarrow \! \widehat{m}^\delta $, and $G^{\delta}_{P}$ learns the reverse mapping $\Psi^{\delta{'}}_{app} \!\!\!\, : m^{\delta} \!\!\!\! \rightarrow \!\! \widehat{p \,}^{\delta} $, where $\widehat{m}^\delta_i$ and $\widehat{p}^\delta_i$ are the generated fake samples; the discriminator $D^{\delta}_{P}\,$/$\,D^{\delta}_{M}$ learn to distinguish real samples $ p^{\delta}   $/$ \, m^{\delta} \,$ and fake samples $  \widehat{p}^\delta \!\!$ / $\! \widehat{m}^\delta\,$.
Our generators $G^{\delta}_{P}$, $G^{\delta}_{M}$ use the Resnet 6 blocks \cite{resnet}, and $D^{\delta}_{P}$, $\,D^{\delta}_{M}$ use the Markovian discriminator of $70\times70$ patchGANs \cite{pixel2pixel, patchgan2, patchgan3}.

%\protect\footnotemark \footnotetext{The detail of network structures as illustrated in our supplementary material.}
We adopt the stable least-squares losses \cite{lsgan} instead of negative log-likelihood objective \cite{gan} as our adversarial losses $L_{adv}$, defined as
 \begin{equation}
\label{equation:Lgan}
\begin{aligned}
\mathcal{L}_{adv}^\delta (  G^{\delta}_{M}, D^{\delta}_{M})   = \ & \mathbb{E}_{m^{\delta} \! \sim \!\, M^{\delta}}[(D^{\delta}_{M}(m^{\delta})-1)^2] \  \\
+ & \  \mathbb{E}_{p^{\delta} \! \sim \, \! P^{\delta}}[D^{\delta}_M(G^{\delta}_M(p^{\delta}))^2]
\end{aligned} \ \ \ ,
\end{equation}
while $\mathcal{L}_{adv}^\delta (  G^{\delta}_{P}, D^{\delta}_{P})$ is defined in a similar manner.

$\mathcal{L}_{cyc}$ is the cycle-consistency loss \cite{cyclegan} that is used to constrain the mapping solution between the input and the output domain, defined as
\begin{equation}
\label{equation:Lcyc}
\begin{aligned}
\mathcal{L}_{cyc}^\delta (G^{\delta}_{P}, G^{\delta}_{M})= \mathbb{E}_{p^{\delta} \! \sim \!\, P^{\delta}}[\| G^{\delta}_P(G^{\delta}_M(\;\! p^{\delta})) - p^{\delta} \|_1 ]\\
+ \ \mathbb{E}_{m^{\delta} \! \sim \!\, M^{\delta}}[\| G^{\delta}_M(G^{\delta}_P(\;\! m^{\delta})) - m^{\delta} \|_1]
\end{aligned} \ \ .
\end{equation}

However, we find that the baseline architectures of ${N^{eye}}$ and ${N^{mouth}}$ with $\mathcal{L}_{adv}$ and $\mathcal{L}_{cyc}$ still fail to preserve the similarity between two domains. Specifically, for regions of eye and mouth, it always produces messy results since the networks almost unable to match colored photos and discrete black lines of mangas. Therefore, we further make three following improvements to optimize their performances.

First, we design a \emph{Similarity Preserving (SP)} module with an SP loss $\mathcal{L}_{S\!P}$ to enhance the similarity. Second, we train an encoder $E^{eye}$ that can extract the main backbone of $p^{eye}$ to binary results, as the input of $N^{eye}$, and an encoder $E^{mouth}$ that encodes $p^{mouth}$ to binary edge-lines, used to guide the shape of manga mouth\footnote{Training details of the two encoders are described in Section \ref{sec:ablation}.}. Third, a structural smoothing loss $\mathcal{L}_{S\!S}$ is designed for encouraging networks to produce manga with smooth stroke-lines, defined as
\vspace{-0.2cm}
\begin{equation}
\label{equation:Lss}
\begin{aligned}
\!\!\!\!\!\!\!\!\!\!\! \mathcal{L}_{S\! S}(G_P^{\delta},\! G_M^{\delta})\!= & {\frac{1}{\sqrt{2\pi}\sigma }} \Big[ \!\!\!\!\!\!\!\!\!\! \!\!\! \sum_{ \ \ \ \ { j\in\{1,2,...,N\}}} \!\!\!\!\!\!\!\!\!\!\!\!\!\!\!  \exp \Big({\frac{-(G_P^{\delta}(m^{\delta})_j-\mu)^2}{2\sigma^2}}\Big) \\[-1mm]
 & + \!\!\!\!\!\!\!\!\! \sum_{{k\in\{1,2,...,N\}}} \!\!\!\!\!\!\!\!\!   \exp\Big({-\frac{(G_M^{\delta}(p^{\delta})_k-\mu)^2}{2\sigma^2}}\Big)\Big]
\end{aligned},
\vspace{-0.1cm}
\end{equation}
where $\mathcal{L}_{S\!S}$ based on a Gaussian model with $\mu$$=$$\frac{255}{2}$, $G_P^{\delta}(m^{\delta})_j$ or $G_M^{\delta}(p^{\delta})_k$ is the $j$-th or $k$-th pixel of $G_P^{\delta}(m^{\delta})$ or $G_M^{\delta}(p^{\delta})$. The underlying idea is that producing unnecessary gray areas will distract and mess the manga results since manga mainly consists of black and white stroke lines. Thus, we give a pixel smaller loss when its gray value closer to black (0) or white (255), to smooth the gradient edges of black stroke lines and produce clean results.

\begin{figure}[t]
\centering
\includegraphics[width=3 in]{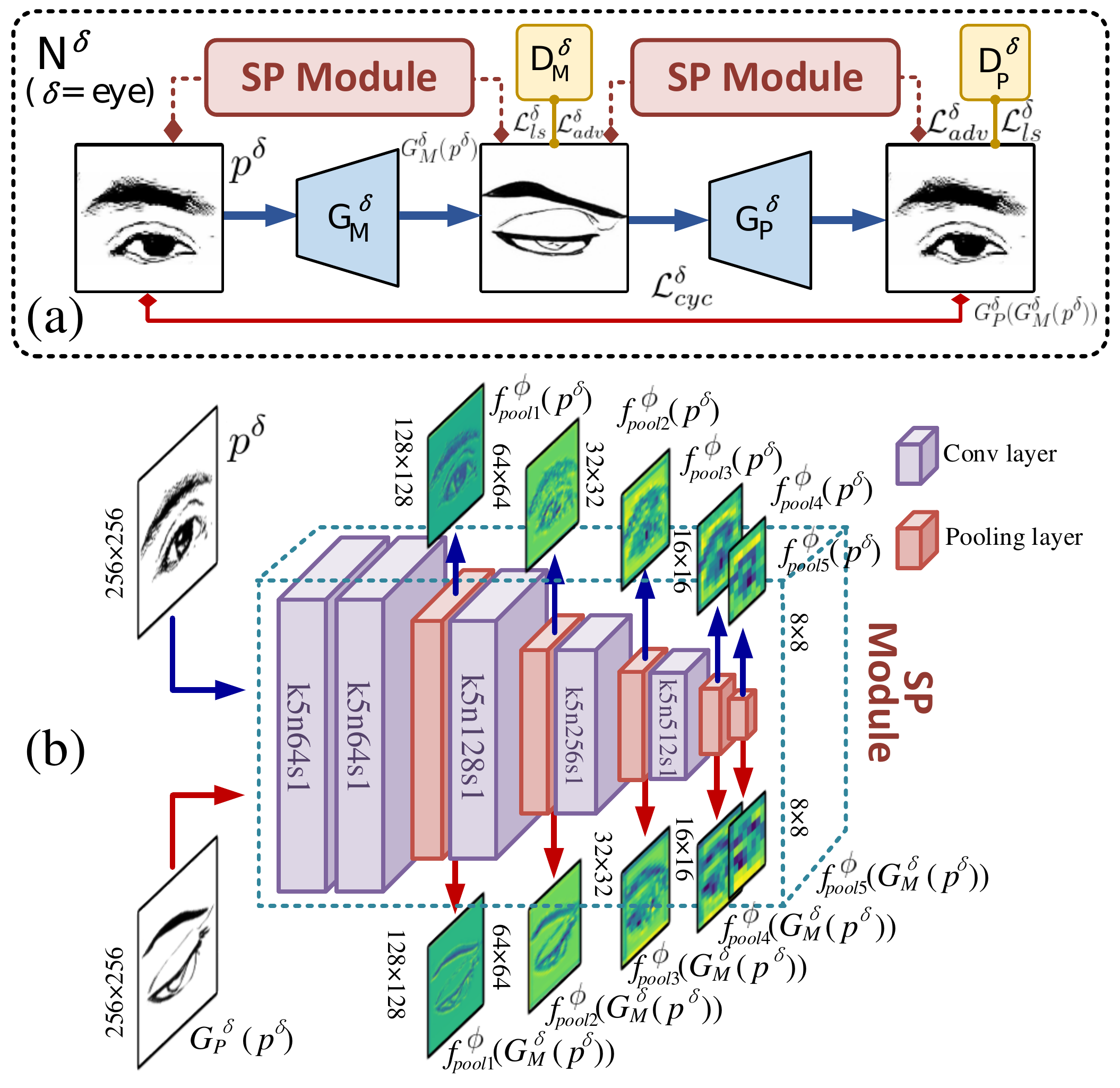}

\caption{(a) We append two SP modules on both forward and backward mappings. (b) SP module extracts feature maps with different resolutions and measures the similarities between two inputs in different latent spaces.}

\label{fig:SP}
\end{figure}

\textbf{Similarity Preserving Module.} The main idea of SP module is that keeping the similarity between two images at a lower resolution can give them similar spatial distributions and different pixel details when they are up-sampled to a higher resolution.
As shown in Figure \ref{fig:SP}(a), we append two SP modules on both forward and backward mappings of $N^{\delta}$. SP module leverages a pre-trained network $\phi $ that we designed to extract feature maps in different latent spaces and resolutions. The architecture of $\phi $ as shown in Figure \ref{fig:SP}(b), it only uses few convolutional layers since we consider the correspondences of encoded features are relatively clear. For the forward mapping $\Psi^{\delta}_{app} \!\! : \widehat{m}^{\delta}\!\!=\!G^{\delta}_M(\;\! p^{\delta})$, we input $p^{\delta}$ and $G^{\delta}_M(\;\! p^{\delta})$ to SP module, and optimize $G^{\delta}_M$ by minimizing the loss functions $\mathcal{L}_{\scriptscriptstyle{S \! P}}(G^{\delta}_M, p^{\delta}  )$ defined as
\vspace{-0.1cm}
\begin{equation}
\label{equation:SP}
\begin{aligned}
\mathcal{L}_{\scriptscriptstyle{S \! P}}(G^{\delta}_M, p^{\delta}  )=  \sum_{i\in \phi}&  \lambda_i \mathcal{L}_{feat}^{\phi,i} \big[ f^{\phi}_i(p^{\delta}),f^{\phi}_i(G^{\delta}_M(\;\! p^{\delta})) \big]\\[-2mm]
& + \ \lambda_{{\scriptscriptstyle{I}}} \mathcal{L}_{pixel}^{{{\scriptscriptstyle{I}}}} \left[\ p^{\delta},G^{\delta}_M(\;\! p^{\delta}) \ \right] \! ,
\end{aligned}
\end{equation}
where $\lambda_i$, $\lambda_{{\scriptscriptstyle{I}}}$ controls the relative importance of each objective, $\mathcal{L}_{pixel}^{{{\scriptscriptstyle{I}}}}$ and $\mathcal{L}_{feat}^{\phi,i}$ are used to keep the similarity on pixel-wise and different feature-wise respectively. $\mathcal{L}_{pixel}^{{{\scriptscriptstyle{I}}}}$ and $\mathcal{L}_{feat}^{\phi,i}$ defined as
\vspace{-0.1cm}
\begin{equation}
\label{equation:fi}
\begin{aligned}
\!\! \!\!\!\!\!\!\! \mathcal{L}_{feat}^{\phi,i}\! \ [ f^{\phi}_i(p^{\delta}),f^{\phi}_i\!(\!G^{\delta}_M(\;\! p^{\delta})) ]&\!\!=\!\!\big\| \! f^{\phi}_i(p^{\delta})\!- \!\! f^{\phi}_i\!(G^{\delta}_M(\;\! p^{\delta})) \;\!\! \big\|_2^2
\\\!\!\! \mathcal{L}_{pixel}^{{{\scriptscriptstyle{I}}}} [\ p^{\delta},G^{\delta}_M(\;\! p^{\delta}) ] & \!\!=\!\! \big\| \! \; p^{\delta}\!\!-\!G^{\delta}_M(\;\! p^{\delta}) \!\;\! \big\|_2^2
\end{aligned}  ,
\end{equation}
where $f^{\phi}_i(x)$ is a feature map extracted from $i$-th layer of network ${\phi}$ when $x$ as the input. Note that we only extract feature maps after pooling layers.

Combining Eq.(\ref{equation:Lgan})-(\ref{equation:fi}), the full objective for learning the appearance mappings of $N^{\delta}$($\delta$$\in$$\{$eye, mouth$\}$) is:
\begin{equation}
\label{equation:total}
\begin{aligned}
\mathcal{L}^\delta _{app}\!  = \! \mathcal{L}_{adv}^\delta  (  G^{\delta}_{M} ,\! D^{\delta}_{M} ) +  \mathcal{L}_{adv}^\delta  ( G^{\delta}_{P}, \! D^{\delta}_{P} ) \\
    + \ \alpha_1 \mathcal{L}_{cyc}^\delta (G^{\delta}_{P}, G^{\delta}_{M})
  +  \alpha_2 \mathcal{L}_{S \! P}^\delta(G^{\delta}_{M}, p^{\delta}) \\
 + \ \alpha_3 \mathcal{L}_{S \! P}^\delta(G^{\delta}_{P}, m^{\delta}) + \alpha_4 \mathcal{L}_{S \! S}(G^{\delta}_{M},G^{\delta}_{P})
\end{aligned}  \ \  ,
\end{equation}
where $\alpha_1$ to $\alpha_4$ used to balance the multiple objectives.
%Thus following the methodology of manga drawing , we design a multi-GAN architecture including three locals GAN for each facial feature, instead of translating a whole image. Although the local GAN has been used in several cross-domin translation works, e.g., APDrawingGAN\cite{apdrawinggan}, TP-GAN \cite{tpgan}, they often use paired data or ground truth to constantly correct the produced results. Contrarily, our unpaired task without any referable ground truth, since we present different tailored strategies in each local network to optimize their performances.

%The three local GAN $G_eye$, $G_nose$, and  to translate the regions of eyes, nose, and mouth, respectively.

\subsubsection{$\!\!\!$Translating regions of nose and hair}
Noses are insignificant to manga faces since almost all characters have a similar nose in the target manga style. Therefore, $N^{nose}$ adopts a generating method instead of a translating one, which follows the architecture of progressive growing GANs \cite{pggan} that can produce a large number of high-quality results similar to training data. As shown in Figure \ref{fig:multigan}(d), we first train a variational autoencoder \cite{VAE} to encode the nose region of the input photo into a feature vector, then make the vector as a seed to generate a default manga nose, and we also allow users to change it according to their preferences.

$N^{hair}$ employs a pre-trained generator of APDdrawingGAN \cite{apdrawinggan} that can produce binary portrait hair with the style similar to manga. In addition, the coordinates of generated portraits can accurately correspond to the input photos. As shown in Figure \ref{fig:multigan}(e), we first extract the rough hair region by a hair segmentation method \cite{hairseg} with a fine-tune of expanding the segmented area, and then remove the extra background area by a portrait segmentation method \cite{PortraitFCN}.

\begin{figure}[t]
\centering
\includegraphics[width=3.2 in]{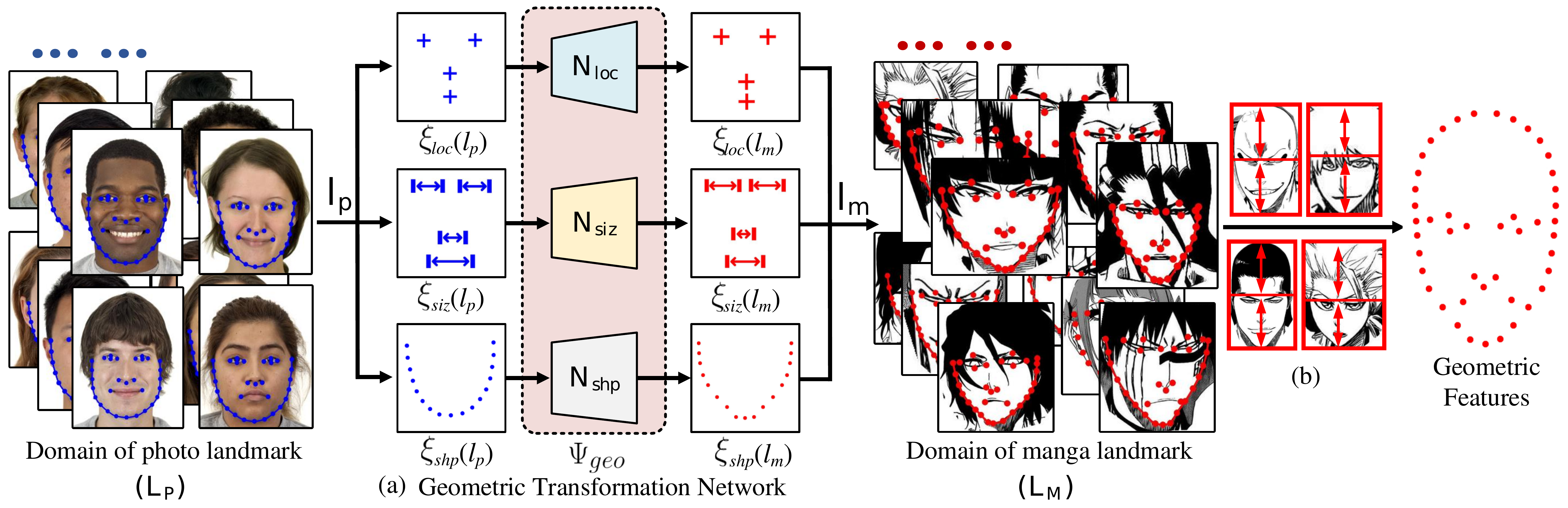}
\vspace{-0.3cm}
\caption{The pipeline of GTN. (a) To improve the variety of facial collocation mode, GTN divides geometric information into three independent attributes, i.e., facial features' locations, sizes, and face shape, while including three sub-GANs $N_{loc}$, $N_{siz}$, $N_{sha}$ to targetedly translate them. (b) According to the pre-computed proportion of cheek and forehead, we produce the geometric features of a whole manga face.}
\vspace{-0.3cm}
\label{fig:GTN}
\end{figure}

\subsection{Geometric transformation network}
\label{section:GTN}
The goal of GTN is to translate the geometric features of faces from the photo domain to the manga domain, where we represent these features with facial landmarks. Let $L_P$ and $L_M$ express the domain of landmarks corresponding to photo and manga. GTN learns a geometric mapping $\Psi_{geo}\!\!: l_p$$\in$$L_P \! \to \! l_m$$\in$$L_M $, where $l_m$ must be similar to $l_p$ and follow manga's geometric style. For training data, each landmark $l_p$ can be extracted by an existing face landmark detector \cite{dlib}, and 106 facial landmarks of manga data $l_m$ are manually marked by us.

 When translating facial landmarks, an issue is that the collocation mode of facial features constrains the variety of results. For example, people with the same face shape may have different sizes or locations of eyes, nose, or mouth. However, GAN may generate them in a fixed or similar collocation mode when it is trained by the landmarks of global faces. Accordingly, as shown in Figure \ref{fig:GTN}, we divide the geometric features into three attributions (face shape, facial features' locations and sizes) and employ three sub-GANs $N_{sha}$, $N_{loc}$, $N_{siz}$ to translate them respectively.

\begin{figure}[t]
\centering
\includegraphics[width=3.2 in]{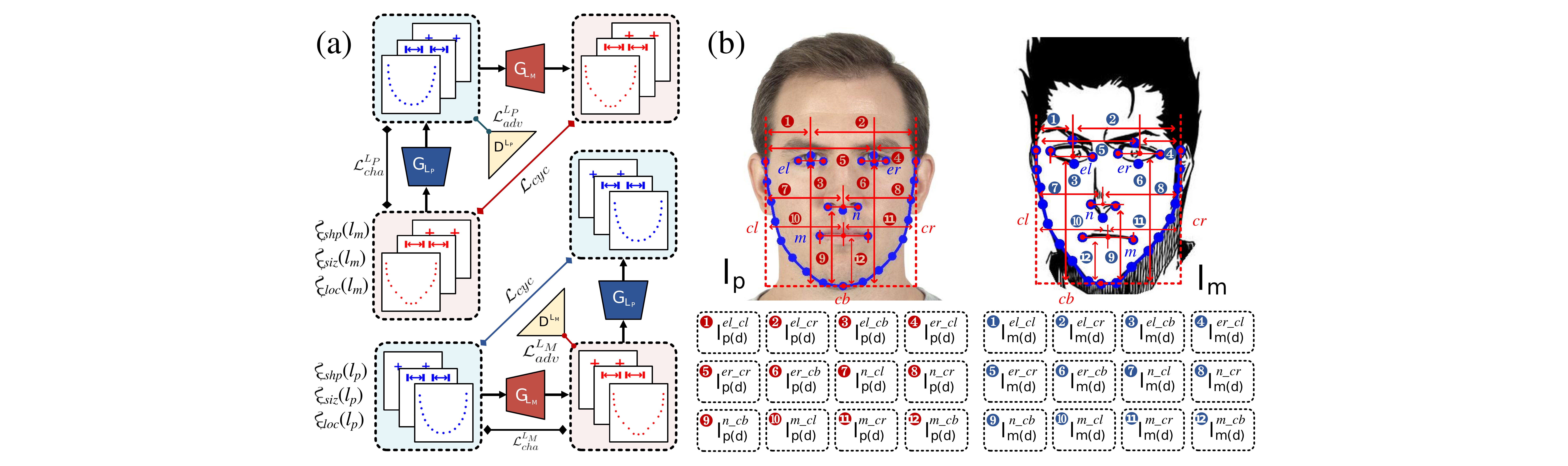}
\caption{(a) Architectures of $N_{loc}$, $N_{siz}$, and $N_{sha}$. (b) Definitions of relative locations in $\xi_{loc}(l_p)$ and $\xi_{loc}(l_m)$.}
\label{fig:GTN2}
\vspace{-0.3cm}
\end{figure}

\textbf{Input of sub-GANs.} For $N_{loc}$, we employ relative locations instead of absolute coordinates, since directly generating coordinates may incur few facial features beyond the face profile. As shown in Figure \ref{fig:GTN2}(b), for $l_p$, relative locations are represented as a vector $\xi_{loc}(l_p)$. $\xi_{loc}(l_p)=\{ l_p^{el}$, $l_p^{er}$, $l_p^{n}$, $l_p^{m} \}$ and $\xi_{loc}(l_m)$ is represented similarly, where $l_p^{el}$, $l_p^{er}$, $l_p^{n}$, $l_p^{m}$ represent regions of left eye, right eye, nose, and mouth respectively. Take $l_p^{n}$ as an example, its relative location is represented as three scalars $l_{p(d)}^{n\_cl}$, $l_{p(d)}^{n\_cr}$, $l_{p(d)}^{n\_cb}$, corresponding to distances of nose's center to cheek's left edge, right edge, and bottom edge respectively, and $l_p^{el}$, $l_p^{er}$, $l_p^{m}$ are defined similarly.
$N_{siz}$ only learns the mapping of facial features' widths, since the length-width ratio of the generated manga facial regions are fixed. Then, the size features of $l_p$ is represent as $\xi_{siz}(l_p)=\{l_{p(w)}^{el}$, $l_{p(w)}^{er}$, $l_{p(w)}^{n}$, $l_{p(w)}^{m}\}$, where $l_{p(w)}^{el}$, $l_{p(w)}^{er}$, $l_{p(w)}^{n}$, $l_{p(w)}^{m}$ represent the width of left eye, right eye, nose, and mouth respectively. $N_{sha}$ learns to translate the face shape, where the face shape is represented as the landmark of cheek region containing 17 points.
%$\xi_{loc}(l_m)$, $\xi_{siz}(l_m)$, and $\xi_{sha}(l_m)$ are defined similar to $\xi_{loc}(l_p)$, $\xi_{siz}(l_p)$, and $\xi_{shp}(l_p)$ respectively.
\begin{figure}[b]
\centering
\includegraphics[width=2.5 in]{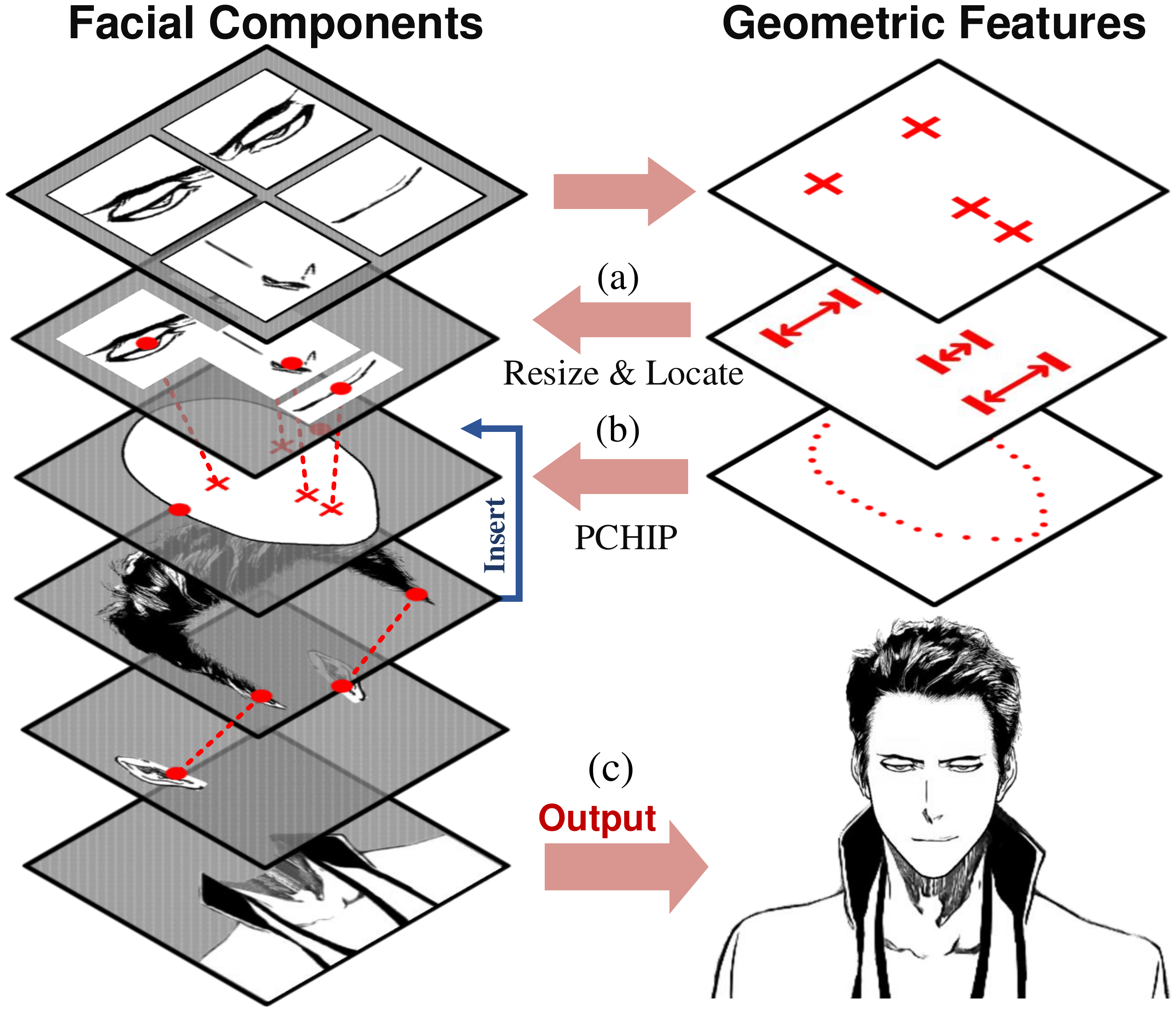}
\vspace{-0.2cm}
\caption{In synthesis module, we generate manga by fusing all facial components and their geometric features.}
\label{fig:SysM}
\vspace{-0.4cm}
\end{figure}

\textbf{Network architecture.} As shown in Figure \ref{fig:GTN2}(a), $N_{loc}$, $N_{siz}$, and $N_{shp}$ roughly follow the structure of CycleGAN \cite{cyclegan} with adversarial loss $\mathcal{L}_{adv}$ as eq(\ref{equation:Lgan}) and cycle loss $\mathcal{L}_{cyc}$ as eq(\ref{equation:Lcyc}). Moreover, we replace all convolutional layer in generators with the fully connected layers, and add the characteristic loss $\mathcal{L}_{cha}$ \cite{carigans} that leverages the differences between a face and the mean face to measure the distinctive features after exaggeration. Let $\mathcal{L}^{\!L_{\!M}}_{cha}(G_{\!L_{\!M}})$ indicates the characteristic loss on the forward mapping, defined as
\begin{equation}
\label{equation:Lcha}
\begin{aligned}
 \mathcal{L}^{\!L_{\!M}}_{cha}(G_{\!L_{\!M}}) =   \mathbb{E}_{\xi_{*}(l_p) \sim \xi_{*}(L_P)} \big\{1 -  \cos[  \xi_{*}(l_p)  \\
    - \  \xi_{*}(\overline{L_{\!P}}),  G_{L_{\!M}}(\xi_{*}(l_p)) - \xi_{*}(\overline{L_{\!M}}) ] \big\}
\end{aligned} \ ,
\end{equation}
where $\xi_{*}(\overline{L_P})$ or $\xi_{*}(\overline{L_M})$ denotes the averages of vector $\xi_{*}({L_P})$ or $\xi_{*}({L_M})$ whose format defined by network $N_{*}$, $*$$ \in$$\{{loc}, {siz}, {shp}\} $, while the reverse loss $\mathcal{L}_{cha}^{L_P}$ is defined similarly.
We let $\mathop{\mathcal{L}}\limits_{\scriptscriptstyle{^{loc}}}$ denotes the loss of $N_{loc}$, and losses of $N_{siz}$ and $N_{sha}$ are represented in a similar manner. The objective function $ \mathcal{L}_{geo}$ to optimize GTN is
\begin{equation}
\label{equation:GTNtotal}
\begin{aligned}
 \mathcal{L}_{geo}= \!\! \mathop{\mathcal{L}}\limits_{\scriptscriptstyle{loc}}\!\!{}^{L_{\!P}}_{adv} \! + \! \mathop{\mathcal{L}}\limits_{\scriptscriptstyle{loc}}\!{}^{L_{\!M}}_{adv} \!  + \! \beta_1 \!\! \mathop{\mathcal{L}}\limits_{\scriptscriptstyle{loc}}\!\!{}_{cyc} \! + \! \beta_2(\mathop{\mathcal{L}}\limits_{\scriptscriptstyle{loc}}\!\!{}^{L_{\!P}}_{cha} \!+ \! \mathop{\mathcal{L}}\limits_{\scriptscriptstyle{loc}}\!\!{}^{L_{\!M}}_{cha})
 \\
  + \! \mathop{\mathcal{L}}\limits_{\scriptscriptstyle{siz}}\!\!{}^{L_{\!P}}_{adv} \! + \! \mathop{\mathcal{L}}\limits_{\scriptscriptstyle{siz}} \!{}^{L_{\!M}}_{adv} \! + \! \beta_3 \!\! \mathop{\mathcal{L}}\limits_{\scriptscriptstyle{siz}}\!\!{}_{cyc} \! + \! \beta_4(\mathop{\mathcal{L}}\limits_{\scriptscriptstyle{siz}}\!\!{}^{L_{\!P}}_{cha} \! + \! \mathop{\mathcal{L}}\limits_{\scriptscriptstyle{siz}}\!\!{}^{L_{\!M}}_{cha})
  \\
  + \!\! \mathop{\mathcal{L}}\limits_{\scriptscriptstyle{sha}}\!\!{}^{L_{\!P}}_{adv} \! + \! \mathop{\mathcal{L}}\limits_{\scriptscriptstyle{sha}}\!\!{}^{L_{\!M}}_{adv} \!  +\beta_5 \!\! \mathop{\mathcal{L}}\limits_{\scriptscriptstyle{sha}}\!\!{}_{cyc} \!+  \! \beta_6 (\mathop{\mathcal{L}}\limits_{\scriptscriptstyle{sha}}\!\!{}^{L_{\!P}}_{cha} \!+\! \mathop{\mathcal{L}}\limits_{\scriptscriptstyle{shp}}\!\!{}^{L_{\!M}}_{cha})
\end{aligned} \ ,
\end{equation}
where $\beta_1$ to $\beta_6$ used to balance the multiple objectives.

Finally, as shown in Figure \ref{fig:GTN}(b), according to the pre-defined proportion of cheek and forehead, we produce the geometric features of the whole manga face.

%\footnotetext{The detail of network structures as illustrated in our supplementary material.}\protect\footnotemark

\begin{figure*}[t]
%\vspace{-0.5cm}
\centering
\includegraphics[width=6.9 in]{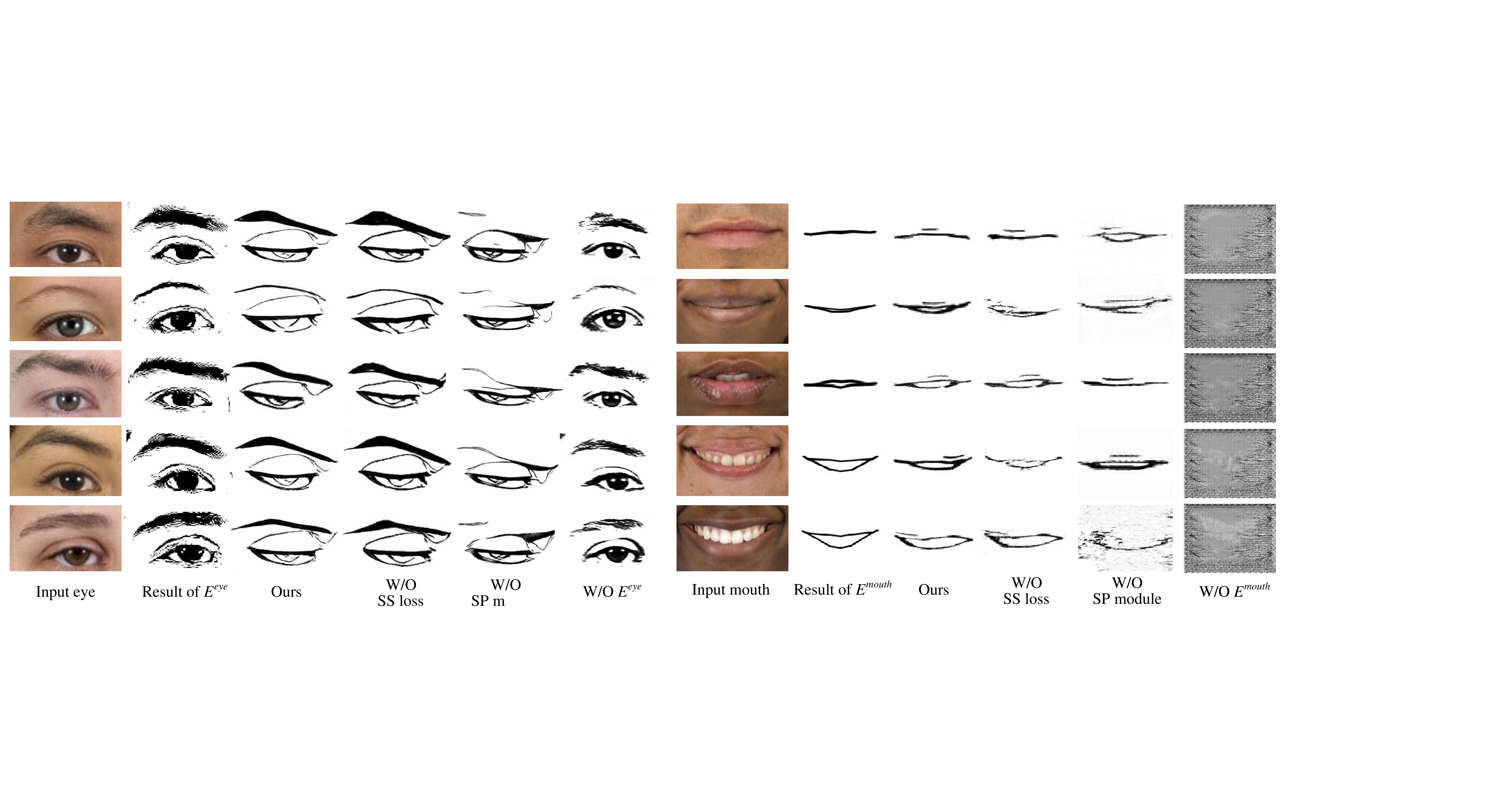}

\caption{Comparison results of eye and mouth regions based on different improvements. Obviously, without our improvements, the network produces poor manga results with messy regions and artifacts, and even cannot capture the correspondences between inputs and outputs.}
\label{fig:without}
\vspace{-0.2cm}
\end{figure*}

\subsection{Synthesis Module}
\label{section:fusion}
The goal of this module is to synthesis an attractive manga face by combining facial components and their geometric features. As mentioned above, facial components of eyes, nose, mouth, and hair are generated by ATN in Section \ref{section:ATN}, and the geometric features of them are generated by GTN in Section \ref{section:GTN}.

The pipeline of fusing components is shown in Figure \ref{fig:SysM}. First, we resize and locate facial components following the geometric features [Figure \ref{fig:SysM}(a)]. Second, the face shape is drawn by the fitting curve of generated landmarks, based on the method of \emph{Piecewise Cubic Hermite Interpolating Polynomial (PCHIP)} \cite{pchip}, where PCHIP can obtain a smooth curve and effectively preserving the face shape [Figure \ref{fig:SysM} (b)]. Then, for ear regions, we provide 10 components of manga ears instead of generating them, since they are stereotyped and unimportant for facial expression. Moreover, we collect 8 manga bodies in our dataset, 5 for male, and 3 for female, that mainly used for decorating faces. In the end, we output a default manga result, and provide a toolkit that allows users to fast fine-tune the size and location of each manga component, and to switch components that insignificant for facial expression (i.e., noses, ears, and bodies) following their preferences [Figure \ref{fig:SysM} (c)].
%\footnote{More details about the synthesis module are illustrated in our supplementary material}

\begin{figure}[t]
%\vspace{-0.5cm}
\centering
\includegraphics[width=3.3 in]{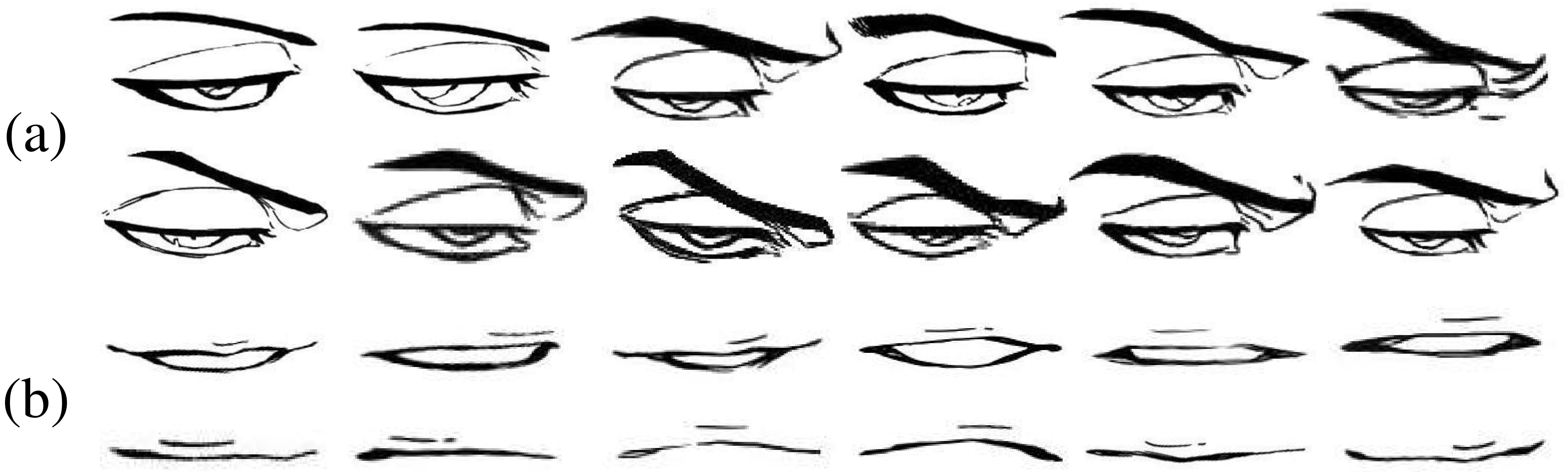}

\caption{(a) Samples of eye regions in target manga work. (b) Samples of mouth regions in target manga work. Comparison with the generated results in Figure \ref{fig:without} and \ref{fig:encode1}, we observe our method effectively preserve the style of the target manga work.}
\label{fig:samples}
\end{figure}

\begin{figure}[t]
%\vspace{-0.2cm}
\centering
\includegraphics[width=3.3 in]{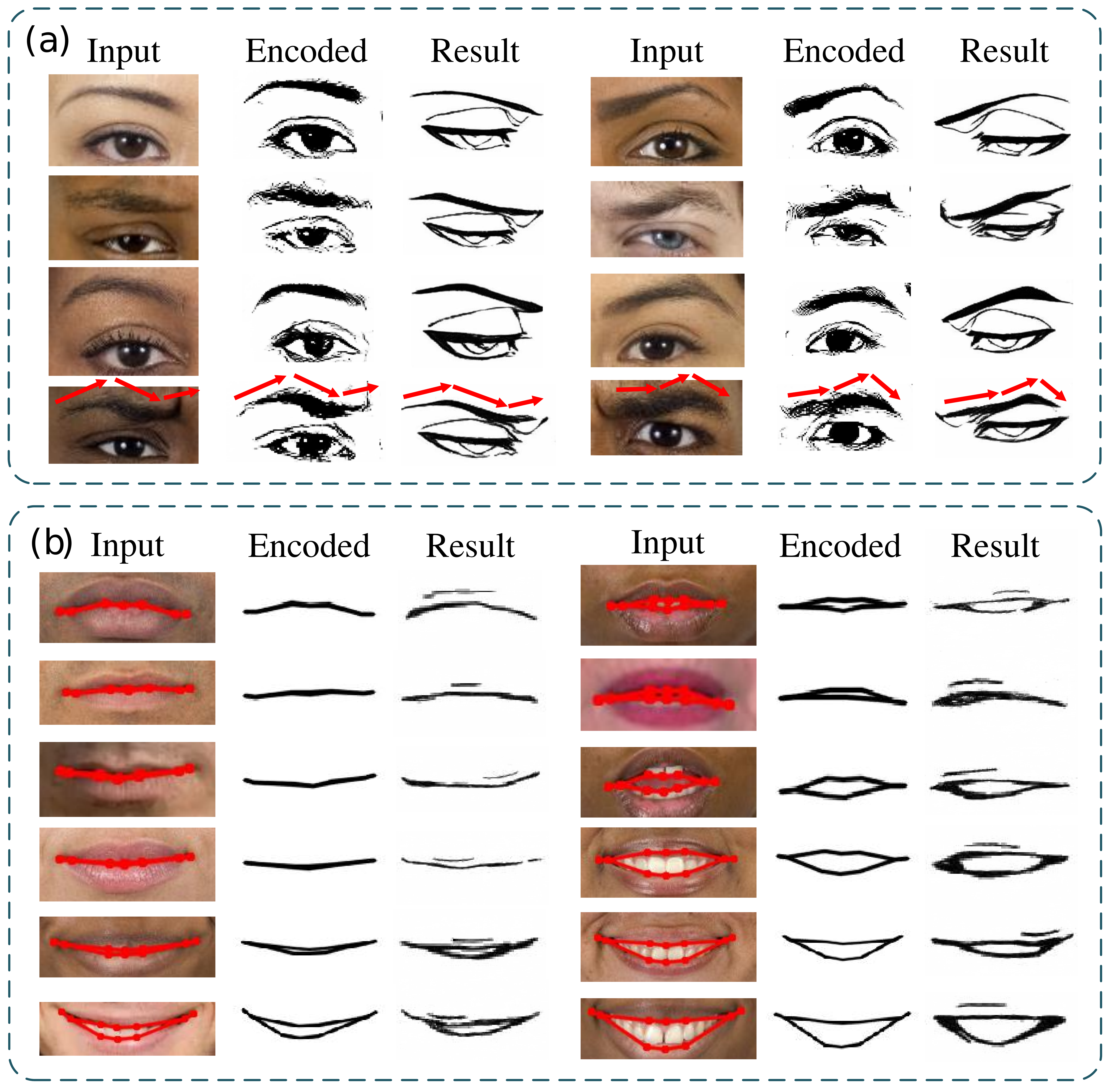}

%\vspace{-0.2cm}
\caption{(a) Samples of eye regions. (b) Samples of mouth regions (red lines indicate landmarks). Our method can effectively preserve the shape of eyebrows (red arrows), eyes, and mouths, and further abstracts them into manga style.}
\label{fig:encode1}
%\vspace{-0.2cm}
\end{figure}

\section{Experiment}
In the following experiments, we first introduce our dataset and training details in Section \ref{sec:data} and then evaluate the effectiveness of our improvements in Section \ref{sec:ablation}. Finally, in Section \ref{sec:compare}, we compare our MangaGAN with other state-of-the-art works. We implemented MangaGAN in PyTorch \cite{pytorch} and all experiments are performed on a computer with an NVIDIA Tesla V100 GPU.

\begin{figure*}[t]
%\vspace{-0.5cm}
\centering
\includegraphics[width=6.8 in]{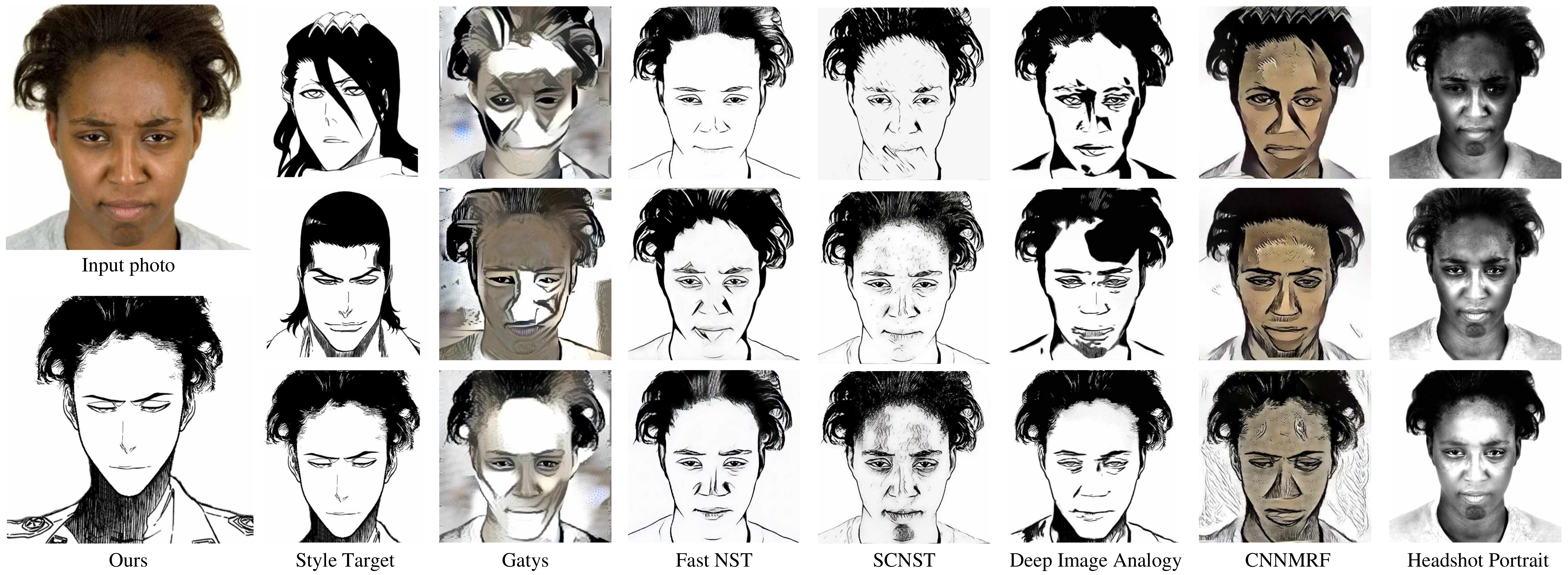}
\caption{Comparison results with NST methods, containing Gatys \cite{Gatys}, Fast NST \cite{Lff}, SCNST \cite{Stroke_NST_ECCV2018}, Deep Image Analogy \cite{dia}, CNNMRF \cite{Markov}, and Headshot Portrait \cite{headshot}. For fair comparison, we employ three different manga faces (one of which is our result) as the style targets to stylize each input photo respectively.}
\label{fig:sotaNST}
%\vspace{-0.2cm}
\end{figure*}

 \subsection{Training}
 \label{sec:data}
 \textbf{Dataset.} The datasets we used in experiments are divided into three parts, i.e., the manga dataset $\mathcal{D}_m$, the photo dataset $\mathcal{D}_p$, and the portrait dataset $\mathcal{D}_b$. $\mathcal{D}_m$, called MangaGAN-BL, is a novel dataset constructed by us and is collected from a world popular manga work \emph{Bleach} \cite{Bleach}. It contains manga facial features of 448 eyes, 109 noses, 179 mouths, and 106 frontal view of manga faces whose landmarks have been marked manually. Moreover, each sample of $\mathcal{D}_m$ is normalized to 256$\times$256 and is optimized by cropping, angle-correction, and repairing of disturbing elements (e.g, covering of hairs, glasses, shadows); $\mathcal{D}_p$ contains 1197 front view of face photos collected from CFD \cite{CFD}, and $\mathcal{D}_b$ contains 1197 black-and-white portraits generated by APDrawingGAN \cite{apdrawinggan} when $\mathcal{D}_p$ as input.

   \textbf{Training details.} For training MangaGAN, each training data of the photo domain and the manga domain is converted to grayscale with 1 channel,
  and each landmark of manga face is pre-processed by symmetric processing to generate more symmetrical faces. For all experiments, we set $\alpha_1$$=$$10$, $\alpha_{\{2,3\}}$$=$$5$, $\alpha_4$$=$$1$ in Eq.(\ref{equation:total}); $\beta_{\{1,3,5\}}$$=$$10$, $\beta_{\{2,4,6\}}$$=$$1$ in Eq.(\ref{equation:GTNtotal}); the parameters of $\mathcal{L}_{S\!P}$ in Eq.(\ref{equation:SP}) are fixed at $\lambda_{{\scriptscriptstyle{I}}}$$=$$1$, $\lambda_{pool5}$$=$$1$, $\lambda_{i}$$=$$0$, $i$$\in$$\{pool1,pool2,pool3,pool4\}$ with the output resolution of 256$\times$256. Moreover, we employ the Adam solver \cite{adam} with a batch size of 5. All networks use the learning rate of 0.0002 for the first 100 epochs, where the rate is linearly decayed to 0 over the next 100 epochs.

\vspace{-0.3cm}

\subsection{Ablation experiment of our improvements}
\label{sec:ablation}
In Section \ref{eyes_and_mouth}, encoders ${E^{eye}}$ and ${E^{mouth}}$ help GANs to capture the abstract correspondences of eye and mouth regions, respectively. $ E^{eye}$ is a conditional GAN model basically following \cite{pixel2pixel}, and is pretrained by paired eye regions of photos from dataset $\mathcal{D}_p$ and their binary result from dataset $\mathcal{D}_b$; $E^{mouth}$ includes a landmark detector \cite{dlib} and a pre-processed program that smoothly connects landmarks of mouth to the black edge-lines to guide the shape of a manga mouth.

With the help of ${E^{eye}}$ and ${E^{mou}}$, as shown in Figure \ref{fig:encode1}, our method can effectively preserve the shape of eyebrows (red arrows), eyes, and mouths, and further abstract them into manga style. Without ${E^{eye}}$ or ${E^{mou}}$, the network cannot capture the correspondences or generated messy results, as shown in the $6^{th}$ and $12^{th}$ columns in Figure \ref{fig:without}.

SP module is essential to keep the similarity between the photo domain and the manga domain. As shown in the $5^{th}$ and $11^{th}$ columns in Figure \ref{fig:without}, without the SP module, neither the manga style nor the similarity between input and output can be well preserved.

Structural Smoothing (SS) loss is also a key to produce mangas with clean appearances and smooth stroke-lines. As shown in the $4^{th}$ and $10^{th}$ columns in Figure \ref{fig:without}, for both eyes and mouth, when training with SS loss, the structure of black stroke lines are effectively smoothed and the gray messy pixels are reduced as well.

\begin{figure*}[t]
\centering
\includegraphics[width=7in]{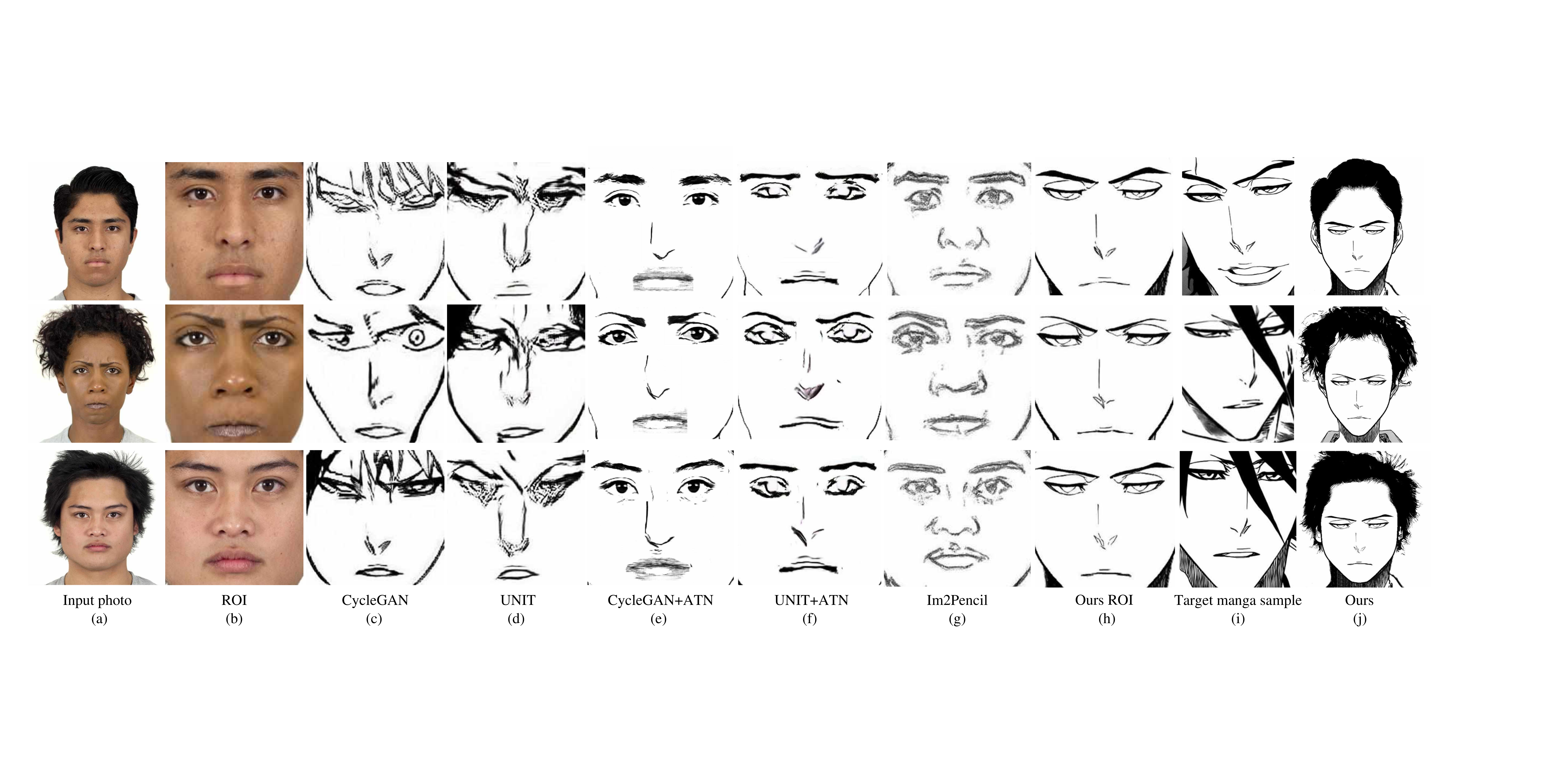}
\vspace{-0.7cm}
\caption{Comparison results with cross-domain translation methods. (a) Input photo. (b) ROI of the input photo. (c)-(h) Results of CycleGAN \cite{cyclegan}, UNIT \cite{UINT}, Im2Pencil \cite{im2pencil}, APDrawingGAN \cite{apdrawinggan}, and our method, respectively. (i) Some typical face samples in target manga work \cite{Bleach}. We obverse that our method can effectively preserve the manga style of (i), e.g., exaggerated eyelid, smooth eyebrow, and simplified mouth. More generated samples as shown in Figure 8 and 9 in our Supplemental Material.}
\label{fig:sotaGAN}
%\vspace{-0.4cm}
\vspace{-0.2cm}
\end{figure*}

\subsection{Comparison with state-of-the-art methods}
\label{sec:compare}
We compare MangaGAN with nine state-of-the-art methods that have potentials to produce manga-like results: the first class is NST methods, containing Gatys \cite{Gatys}, Fast NST \cite{Lff}, SCNST \cite{Stroke_NST_ECCV2018}, Deep Image Analogy \cite{dia}, CNNMRF \cite{Markov}, and Headshot Portrait \cite{headshot}. For fair comparison, as shown in Figure \ref{fig:sotaNST}, we employ three different manga faces (one of which is our result) as the style targets to stylize each input photo respectively. The results show that these methods generally produce warping stroke lines and fail to produce clean manga face, since they focus on transferring the texture and color from the style target. They roughly follow the structure of the photo content, and ignore the transformation of geometric features.

The second class we compared with is cross-domain translation methods, containing CycleGAN \cite{cyclegan}, UNIT \cite{UINT}, and Im2pencil \cite{im2pencil} as shown in Figure \ref{fig:sotaGAN}. For fair comparison, we train CycleGAN and UNIT to translate the whole face region, and translate each facial feature, respectively. For the whole face region translation as shown in Figure \ref{fig:sotaGAN}(c)(d), we only train the ROI [Figure \ref{fig:sotaGAN}(b)] to make these methods easier to find the correspondences between photo and manga, where the photo domain trained by 1197 frontal facial photos' ROIs in $\mathcal{D}_b$, and the manga domain trained by 83 frontal manga faces' ROIs in $\mathcal{D}_m$. For each facial feature translation as shown in Figure \ref{fig:sotaGAN}(e)(f), we append CycleGAN and UNIT on the ATN structure, and train each facial region by the same data as we use.
Comparison results in Figure \ref{fig:sotaGAN} show that the other methods get trouble in matching the poor correspondences between photo and manga, i.e., they focus on matching the dark region of photos and manga, and do not translate the face shape and stroke-line structures. Unlike them, our method can effectively make the output similar to the appearance of the target manga (e.g., exaggerated eyelids, smooth eyebrows, simplified mouths) as shown in Figure \ref{fig:sotaGAN}(h)(i).

%

%For more fair comparisons, we train these 11 advanced methods using the same training data like ours, to translate the same local facial regions, e.g., eye and mouth. Moreover, we also conduct a user study to evaluate the performances of our MangaGAN on subjectively. These experimental results are shown in our supplementary materials.

\section{Discussion}
%Different from some existing advanced methods of which goal is to preserve identity information as much as possible,
\textbf{The performance on preserving manga style}.
Most of the state-of-the-art methods prone to translate the color or texture of the artistic image, and ignore the translation of geometric abstraction. As shown in Figure \ref{fig:sotaNST} and \ref{fig:sotaGAN}, the stylized faces they generated are similar to the input photos with only color or texture changing, which makes them more like the realistic sketches or portraits than the abstract mangas. Unlike them, we extend the translation to the structure of stroke lines and the geometric abstraction of facial features (e.g., simplified eyes and mouths, beautified facial proportions), which makes our results more like the works drawn by the manga artist.
%realistic sketches or portraits than the abstract manga.

%Most of the state-of-the-art methods prone to only translate the color or texture of the artistic image, and typically ignore the translation of geometric proportion and exaggerated shape. Thus, as shown in Figure \ref{fig:sotaNST} and \ref{fig:sotaGAN}, the main structures of the manga faces they generated are corresponding to the input photos, which makes their results almost the same as the input, and more like the
%realistic sketches or portraits than the abstract manga. Different from them, we extend manga style to the structure of stroke lines and the geometric abstraction of facial features (e.g., simplified eyes and mouths, facial proportions), which makes our results more like the works drawn by the target manga artist.
 %manga simplifies and beautifies a human face, i.e.,
\textbf{The performance on preserving user identity}.
We generate manga face guided by the input photo, however, manga characters are typically fictitious, simplified, idealized and much unlike real people. Specifically, manga faces are usually designed to own optimum proportions, and the facial features are simplified to several black lines [Figure \ref{fig:sotaGAN}(i)]. Therefore, the excessive similarity between the output and input will make the output unlike a manga. To generate typical and clean manga faces, we even remove the detail textures and beautify the proportions of facial features, which compromise the performance on preserving the user identity. Accordingly, it is reasonable that there are some dissimilarities between the output manga face and the input facial photo.
%Our goal is to generate a typical manga face, with preserving both the target manga style and the facial similarity captured from the input photo.

\textbf{More evaluations}. To subjectively evaluate the performances of our methods on preserving manga style, user identity, and visual attractiveness, we conduct a series of user studies in Section 2 of the supplementary materials. Moreover, we also show more experimental results and generated manga faces in Section 5 of our supplementary materials.

\vspace{-0.2cm}
\section{Conclusion}
In this paper, we propose the first GAN-based method for unpaired photo-to-manga translation, called MangaGAN. It is inspired by the prior-knowledge of drawing manga, and can translate a frontal face photo into the manga domain with preserving the style of a popular manga work. Extensive experiments and user studies show that MangaGAN can produce high-quality manga faces and outperforms other state-of-the-art methods.

\newpage

\section{Supplemental Material}
\subsection{Overview}
In this document we provide the following supplementary contents:
\begin{itemize}\setlength{\itemsep}{0pt}
\item a series of user studies to subjectively evaluate our method and related state-of-the-art works (Section \ref{sec:userstudy});
\item more details about the ablation experiment of our improvements (Section \ref{sec:ablation});
\item more qualitative results of comparison with state-of-the-art style methods (Section \ref{sec:compare});
\item details about our network architectures (Section \ref{sec:network});

\item more generated samples of our MangaGAN (Section \ref{sec:results});
\item our dataset and download link (Section \ref{sec:data});
\item some failure cases (Section \ref{sec:failcase}).
\end{itemize}

\subsection{User Study}
\label{sec:userstudy}
To subjectively evaluate our performances on preserving manga style, user identity and visual attractiveness, we conduct two user studies in Section \ref{sec:style} and Section \ref{sec:atrac} respectively.
\begin{figure}[b]
\centering
\includegraphics[width=2.5 in]{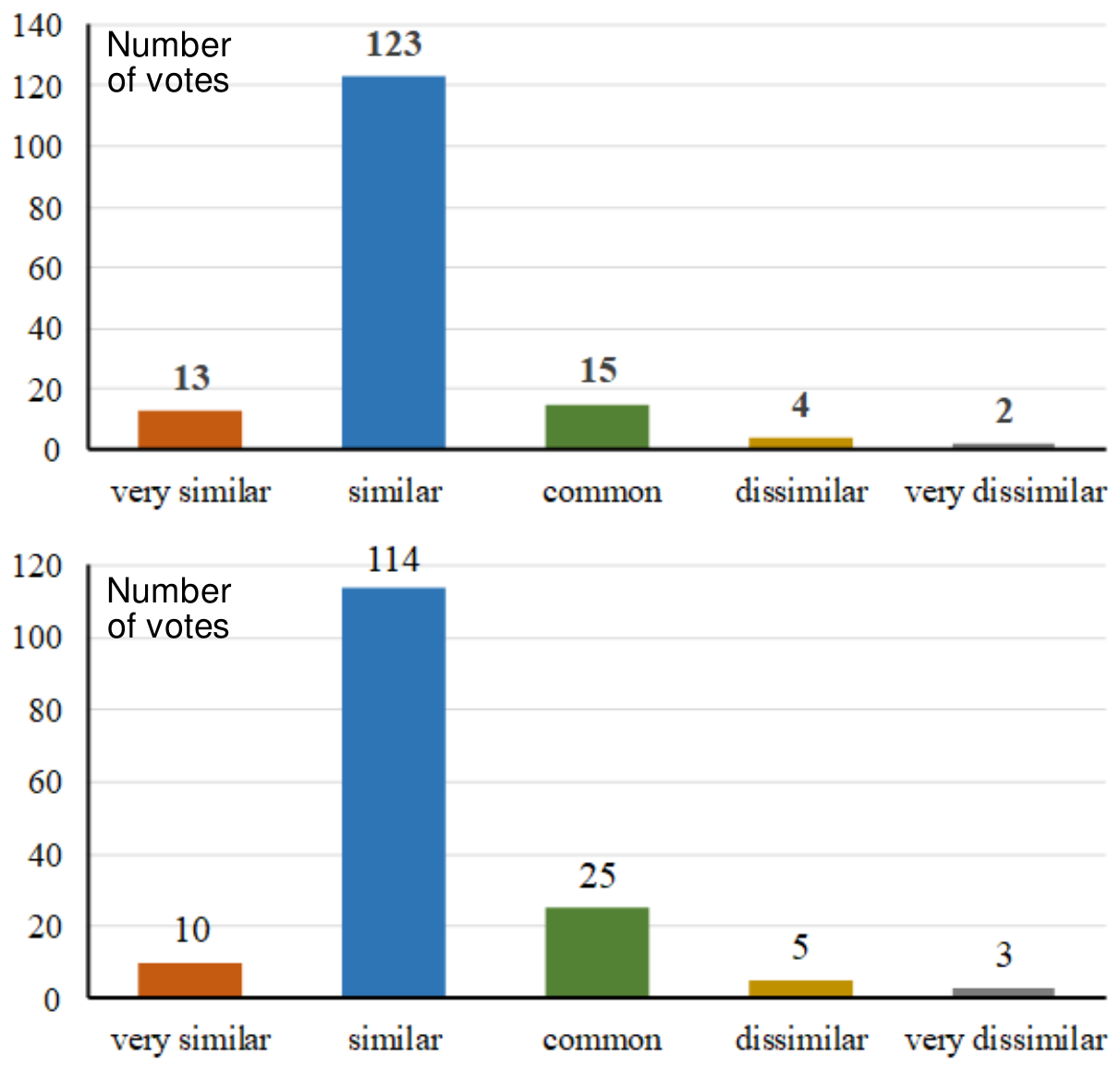}
\caption{Results of our online user study. \emph{Upper:} The user study on how much the similarity between our results and target manga style. \emph{Bottom:} The user study on how much the similarity between our results and input photos.}
\label{fig:userstudy1}
\end{figure}

\subsection{$\!\!\!$Preserve manga style and user identity}
\label{sec:style}
\textbf{Method.} We design an online questionnaire, which first shows some samples of input photo and our corresponding results, and then appends two questions, ``\emph{How much do you think our results are similar to the target manga style?}'' and ``\emph{How much do you think our results are similar to the input photos?}''. All users are required to vote one of five selections (very dissimilar, dissimilar, common, similar, and very similar) according to their observation. To evaluate our work professionally, we anonymously open the questionnaire to a professional manga forum, and ask the experienced manga readers to attend this user study.

\begin{figure}[t]
\centering
\includegraphics[width=3.3 in]{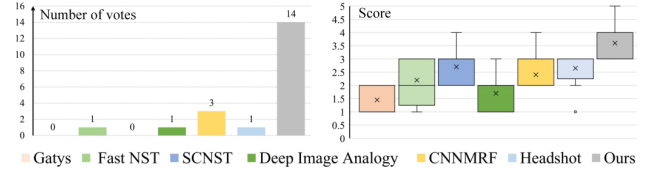}
\caption{User studies on the visual attractiveness of NST methods and ours. \emph{Left}: Voting results of the method that has the most attractive results. \emph{Right}: Boxplot of scoring results for visual attractiveness.}
\label{fig:userstudy}
\end{figure}

\textbf{Result.} In a two-week period, 157 participants attended this user study. The summarized results as shown in Figure~\ref{fig:userstudy}. We observe that 86.62 $\%$ participants believe our results preserve the style of target manga, and 78.98 $\%$ participants believe our results are similar to the input photos, which indicates that our method has good performances on both two aspects.

%\textbf{Discussion.}
%Manga characters are fictitious, abstract, and very different from real people. Specifically, manga typically simplifies and beautifies a human face (e.g., manga face owns a perfect proportion, mouth and nose are simplified to several black lines). Therefore, if the output is too similar to the input photo may make it does not look like a manga. To generate typical and clean manga faces, our method beautifies the face shape and facial features following the proportions of manga faces and removes the detail texture, which may compromise the performance on preserving the person identity. Accordingly, it is reasonable to exist some dissimilarities between the generated manga face and the input photo.
%\begin{figure}[t]
%\centering
%\includegraphics[width=3 in]{Supmat_dis.pdf}
%\caption{(a) Some typical face samples in target manga work \cite{bleach}. (b) Inputs and outputs of our method.}
%\label{fig:dis}
%\end{figure}
\begin{figure}[t]
\centering
\includegraphics[width=3.3 in]{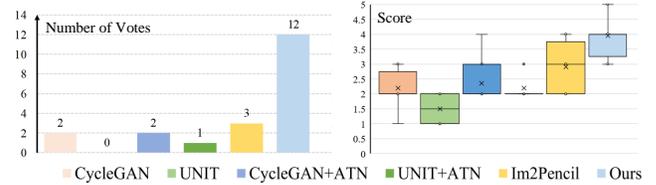}
\caption{User studies on the visual attractiveness of cross-domain translation methods and ours. \emph{Left}: Voting results of the method that has the most attractive results. \emph{Right}: Boxplot of scoring results for visual attractiveness.}
\label{fig:userstudy2}
\vspace{-0.5cm}
\end{figure}

\subsection{Visual attractiveness}
\label{sec:atrac}

\textbf{Method.} We invited 20 volunteers (10 males and 10 females) irrelevant to this work to conduct a user study. Preparing for the experiment, we firstly select ten face photos from $\mathcal{D}_p$ randomly. Then, each photo is expanded to two group of images. The first group containing: one input photo and stylized results that are produced by six NST methods (Gatys \cite{Gatys}, Fast NST \cite{Lff}, SCNST \cite{Stroke_NST_ECCV2018}, Deep Image Analogy \cite{dia}, CNNMRF \cite{Markov}, Headshot Portrait \cite{headshot}), and our MangaGAN, respectively; another group containing: one input photo and stylized results that are produced by five cross-domain translation methods (CycleGAN \cite{cyclegan}, UNIT \cite{UINT}, CycleGAN+ATN, UNIT+ATN, and Im2pencil \cite{im2pencil}), and our MangaGAN, respectively.
    Finally, each volunteer is asked to complete two tasks for each image group: the first task is scoring 1 to 5 for each method's result, where a higher score indicates a higher attractiveness; another task is to vote for the method with the most attractive results.

\textbf{Result.} As shown in Figure~\ref{fig:userstudy}, compared with NST methods, our method scored the highest on visual-quality, and over $70\%$ volunteers believe our results are the most attractive ones. As shown in Figure~\ref{fig:userstudy2}, compared with cross-domain translation methods, our method still gets the highest score and the most number of votes. The above user studies show that our MangaGAN has reached the state-of-the-art level on visual attractiveness.

\subsection{Supplemental Experiment}
\subsubsection{Ablation experiment of our improvements}
\label{sec:ablation}
\begin{figure*}[htbp]
%\vspace{-1cm}
\centering
\includegraphics[width=6.8 in]{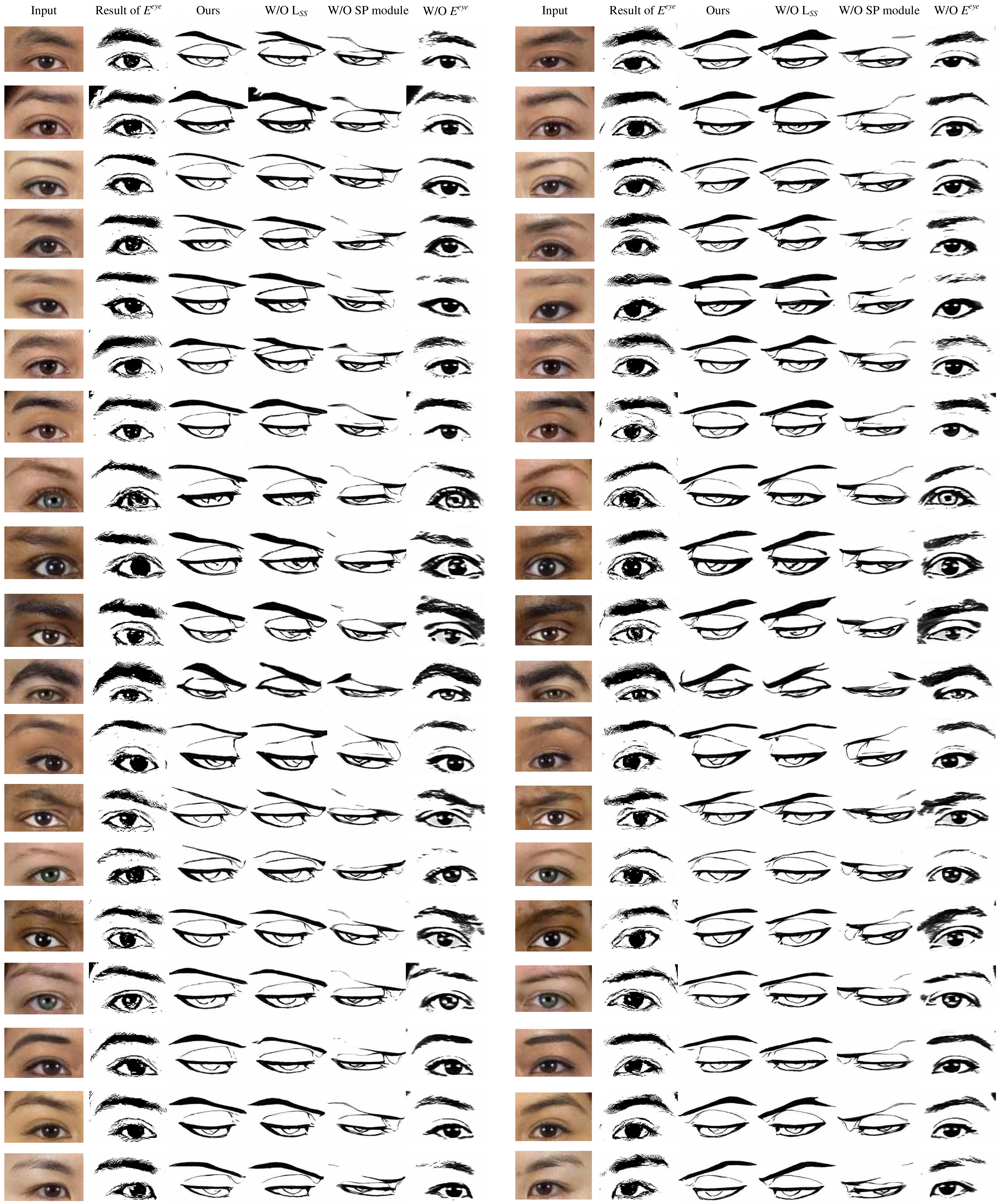}
\caption{Ablation experiment of our improvements on eye regions. \emph{From left to right}: input face photos, results of encoder $E^{eye}$, our results, results of removing structural smoothing loss $L_{S\!S}$, results of removing SP module, and results of removing $E^{eye}$.}
\label{fig:without_supp}
%\vspace{-1cm}
\end{figure*}

In Figure \ref{fig:without_supp}, we show more comparison results corresponding to the ablation experiments in Section {4.2} of the main paper. We can observe that: the structural smoothing
loss $L_{S\!S}$ can make the structure of stroke lines smooth, and constrain the generation of mess gray areas; the SP module successfully preserves the similarity between the input photos and the output mangas; the encoder $E^{eye}$ effectively helps the network extract the main structure of the eye region and capture the poor correspondences between photos and mangas. Without the above improvements, the model cannot generate high-quality results with clean stroke lines and an attractive manga style.

\subsubsection{More qualitative results of comparison}
\label{sec:compare}
According to Section {4.3} of the main paper, for more fair comparisons, we leverage related state-of-the-art methods and our methods to translate the same local facial regions (e.g., eye and mouth) respectively. For NST methods, we use three different manga eyes and mouths (one of which is our result) as the style targets to stylize the input photo respectively. For cross-domain translation methods, we train them to translate the same local facial region, using the same dataset as us.

\begin{figure*}[htbp]
%\vspace{-1cm}
\centering
\includegraphics[width=6.3 in]{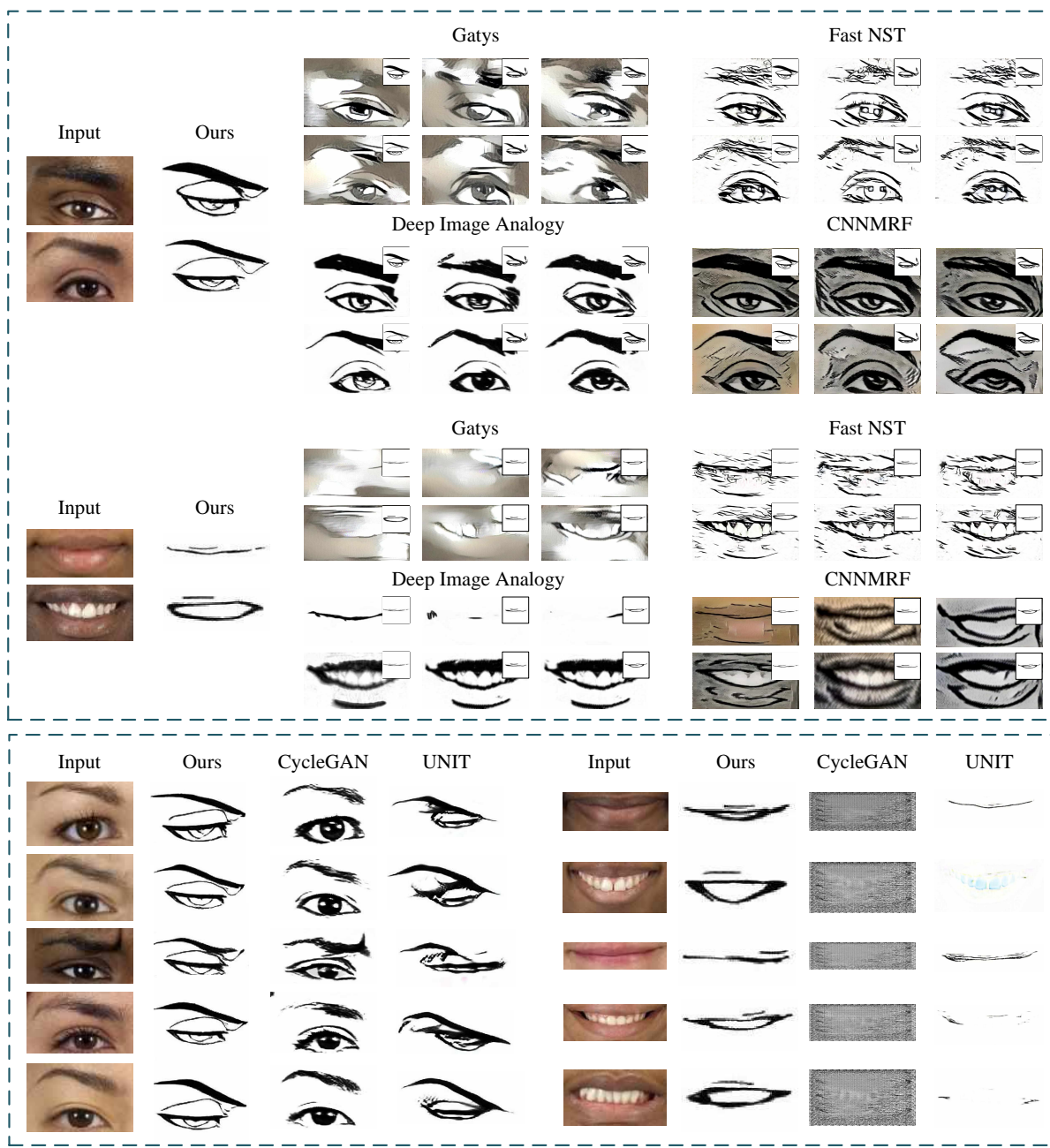}
\caption{\emph{Upper:} comparison results with NST methods, containing Gatys \cite{Gatys}, Fast NST \cite{Lff}, Deep Image Analogy \cite{dia}, and CNNMRF \cite{Markov}. \emph{Bottom:} comparison results with GAN-based one-to-one translation methods, containing CycleGAN \cite{cyclegan} and UNIT \cite{UINT}.}
\label{fig:compare}
\vspace{-0.5cm}
\end{figure*}

Comparison results as shown in Figure \ref{fig:compare}. We observe that neither the NST methods nor the cross-domain methods can generate clean and attractive manga eyes and mouthes, due to the reasons we concluded in Section {4.3} of the main paper.

%Besides, our method harbors a good performance on preserving the similarity between the input photo and the produced manga. Of course, it is impractical to generate exactly similar manga face, since the style of manga is always the abstraction and beautification of real things, where too much similarity will reduce their attractive.

\subsection{Network Architecture}
\label{sec:network}
In Section {3.2} of the main paper, $N^{eye}$, $N^{mouth}$, and $N_{nose}$ are respectively trained for translating facial regions of eye, mouth, and nose, from the input photo $p\in P$ to the output manga $m \in M$. The generators of $N^{eye}$ and $N^{mouth}$ use the Resnet 6 blocks \cite{resnet,cyclegan}, and the discriminators use the Markovian discriminator of $70\times70$ patchGANs \cite{pixel2pixel, patchgan2, patchgan3}. We also tested using U-Net \cite{unet} or Resnet 9 blocks \cite{resnet} as the generators of $N^{eye}$ and $N^{mouth}$, but they often produce messy results. Table \ref{tab:Neye} illustrates the network architectures used for the generators of $N^{eye}$ and $N^{mouth}$.
\renewcommand\arraystretch{1.3}
\begin{table}[h]\scriptsize
\center
\caption{Network architecture used for the generators of $N^{eye}$ and $N^{mouth}$.}
\begin{tabular}{c|c|c|c}
\hline\hline
{Type} & {Kernal Size} & {Output Channels} & {Output Size} \\ \hline\hline
 Input    &    N/A         & 1               & 256          \\ \hline
 Conv    &   7          & 64              & 256         \\ \hline
 ReLu+Conv+IN   &    3         & 128             & 128         \\ \hline
 Residual block   &    3         & 256             & 64         \\ \hline
  Residual block   &    3         & 256             & 64         \\ \hline
   Residual block   &    3         & 256             & 64         \\ \hline
    Residual block   &    3         & 256             & 64         \\ \hline
     Residual block   &    3         & 256             & 64         \\ \hline
      Residual block   &    3         & 256             & 64         \\ \hline

 ReLu+DeConv+IN &    3       & 128           & 128         \\ \hline
 ReLu+DeConv+IN &    3       & 64           & 256         \\ \hline
 ReLu+Conv+IN &    7       & 1           & 256         \\ \hline
\end{tabular}
\label{tab:Neye}
\end{table}

$N^{nose}$ employs a generating method instead of a translating one, which follows the architecture of progressive growing GANs \cite{pggan}. The network architectures of $N^{nose}$ as illustrated in Table \ref{tab:Nnose}.
\renewcommand\arraystretch{1}
\begin{table}[h]\scriptsize
\center
\caption{Network architecture used for the generators of $N^{nose}$.}
\begin{tabular}{c|c|c|c}
\hline\hline
\multicolumn{4}{c}{Generator} \\ \hline
{Type} & {Kernal Size} & {Output Channels} & {Output Size} \\ \hline\hline
Latent vector    &    N/A         & 512               & 1          \\
Conv+LReLU    &   4          & 512              & 4         \\
Conv+LReLU    &   3          & 512              & 4         \\ \hline
Upsample    &    N/A         & 512               & 8          \\
Conv+LReLU    &   3          & 512              & 8         \\
Conv+LReLU    &   3          & 512              & 8         \\ \hline
Upsample    &    N/A         & 512               & 16          \\
Conv+LReLU    &   3          & 512              & 16         \\
Conv+LReLU    &   3          & 512              & 16         \\ \hline
Upsample    &    N/A         & 512               & 32          \\
Conv+LReLU    &   3          & 512              & 32         \\
Conv+LReLU    &   3          & 512              & 32         \\ \hline
Upsample    &    N/A         & 512               & 64          \\
Conv+LReLU    &   3          & 256              & 64         \\
Conv+LReLU    &   3          & 256              & 64         \\ \hline
Upsample    &    N/A         & 256               & 128          \\
Conv+LReLU    &   3          & 128              & 128         \\
Conv+LReLU    &   3          & 128              & 128         \\ \hline
Upsample    &    N/A         & 64               & 256          \\
Conv+LReLU    &   3          & 64              & 256         \\
Conv+LReLU    &   3          & 64              & 256         \\
Conv+liner    &   1         & 3              & 256         \\ \hline\hline

\multicolumn{4}{c}{Discriminator} \\ \hline
{Type} & {Kernal Size} & {Output Channels} & {Output Size} \\ \hline
Input image   &    N/A         & 3              & 256          \\
Conv+LReLU    &   1          & 64              & 256         \\
Conv+LReLU    &   3          & 64              & 256         \\
Conv+LReLU    &   3          & 128              & 256         \\
Downsample    &    N/A         & 128               & 128          \\\hline
Conv+LReLU    &   3          & 128             & 128        \\
Conv+LReLU    &   3          & 256              & 128         \\
Downsample    &    N/A         & 256               & 64          \\\hline
Conv+LReLU    &   3          & 256             & 64        \\
Conv+LReLU    &   3          & 512              & 64         \\
Downsample    &    N/A         & 512               & 32         \\\hline
Conv+LReLU    &   3          & 512             & 32        \\
Conv+LReLU    &   3          & 512              & 32         \\
Downsample    &    N/A         & 512               & 16          \\\hline
Conv+LReLU    &   3          & 512             & 16        \\
Conv+LReLU    &   3          & 512              & 16         \\
Downsample    &    N/A         & 512               & 8          \\\hline
Conv+LReLU    &   3          & 512             & 8        \\
Conv+LReLU    &   3          & 512              & 8         \\
Downsample    &    N/A         & 512               & 4          \\\hline
Conv+LReLU    &   3          & 512             & 4        \\
Conv+LReLU    &   3          & 512              & 4         \\
Conv+LReLU    &   4          & 512              & 1         \\
Fully-connected+linear    &    N/A         & 1               & 1          \\\hline

\end{tabular}
\label{tab:Nnose}
\end{table}

%\begin{figure}[t]
%\centering
%\includegraphics[width=3 in]{data.pdf}
%\caption{ Some samples in dataset MangaGAN-BL, which are optimized by cropping, angle-correcting, and repairing of disturbing elements (e.g, covering of hairs, glasses, shadows). }
%\label{fig:data}
%\end{figure}

\emph{MangaGAN-BL} can be downloaded by the \textbf{\emph{Google Drive}} link: \url{https://drive.google.com/drive/folders/1viLG8fbT4lVXAwrYBOxVLoJrS2ZTUC3o?usp=sharing}.

\begin{table*}[htbp]
\centering
\caption{Some samples of eye regions in input photos and generated mangas. }
\begin{tabular}{c}
\centering
\begin{minipage}{1\textwidth}
\includegraphics[width=17cm]{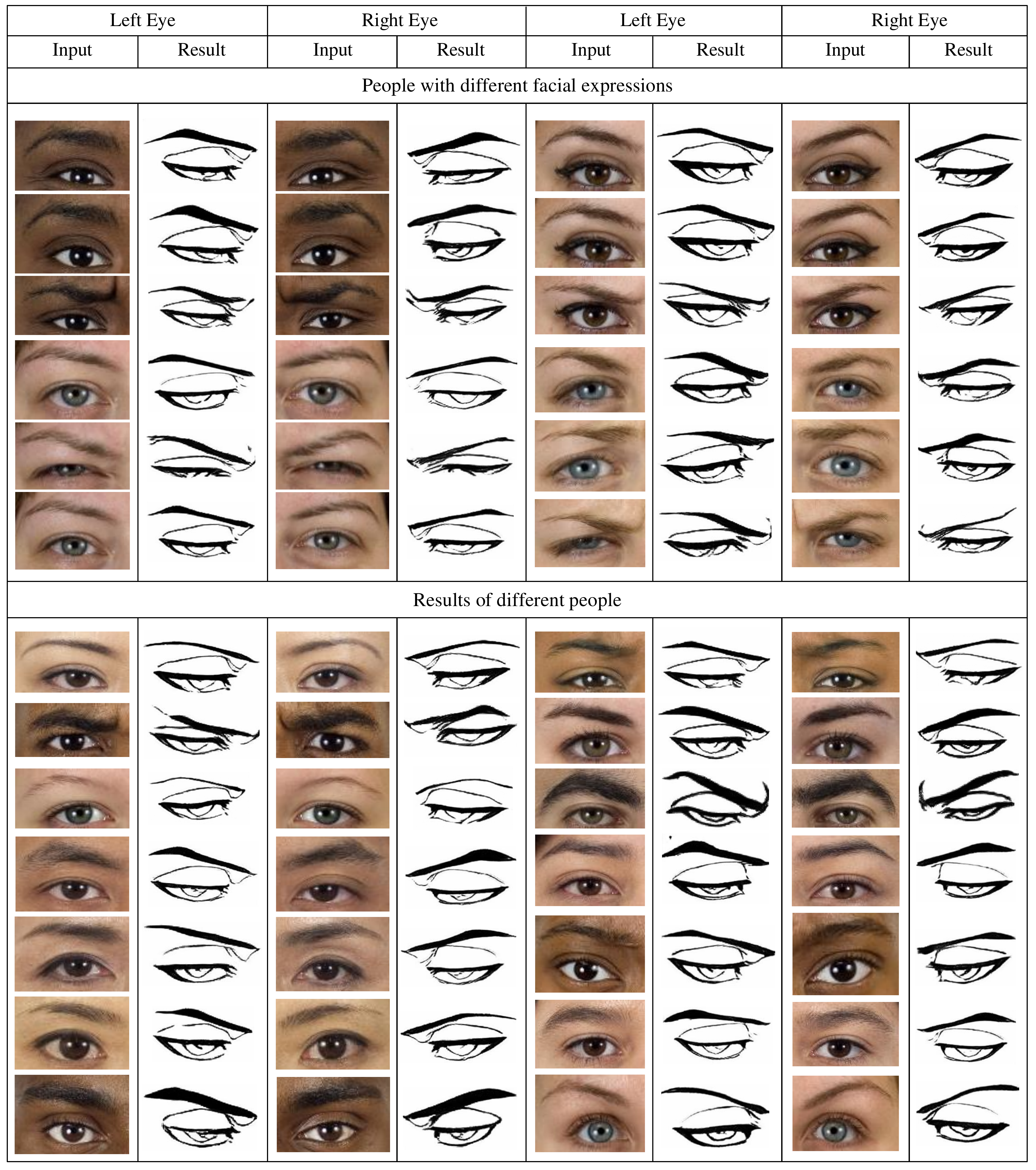}
\end{minipage}
\end{tabular}
\label{tab:eye}
\end{table*}

\begin{figure*}[htbp]
\centering
\includegraphics[width=6.9 in]{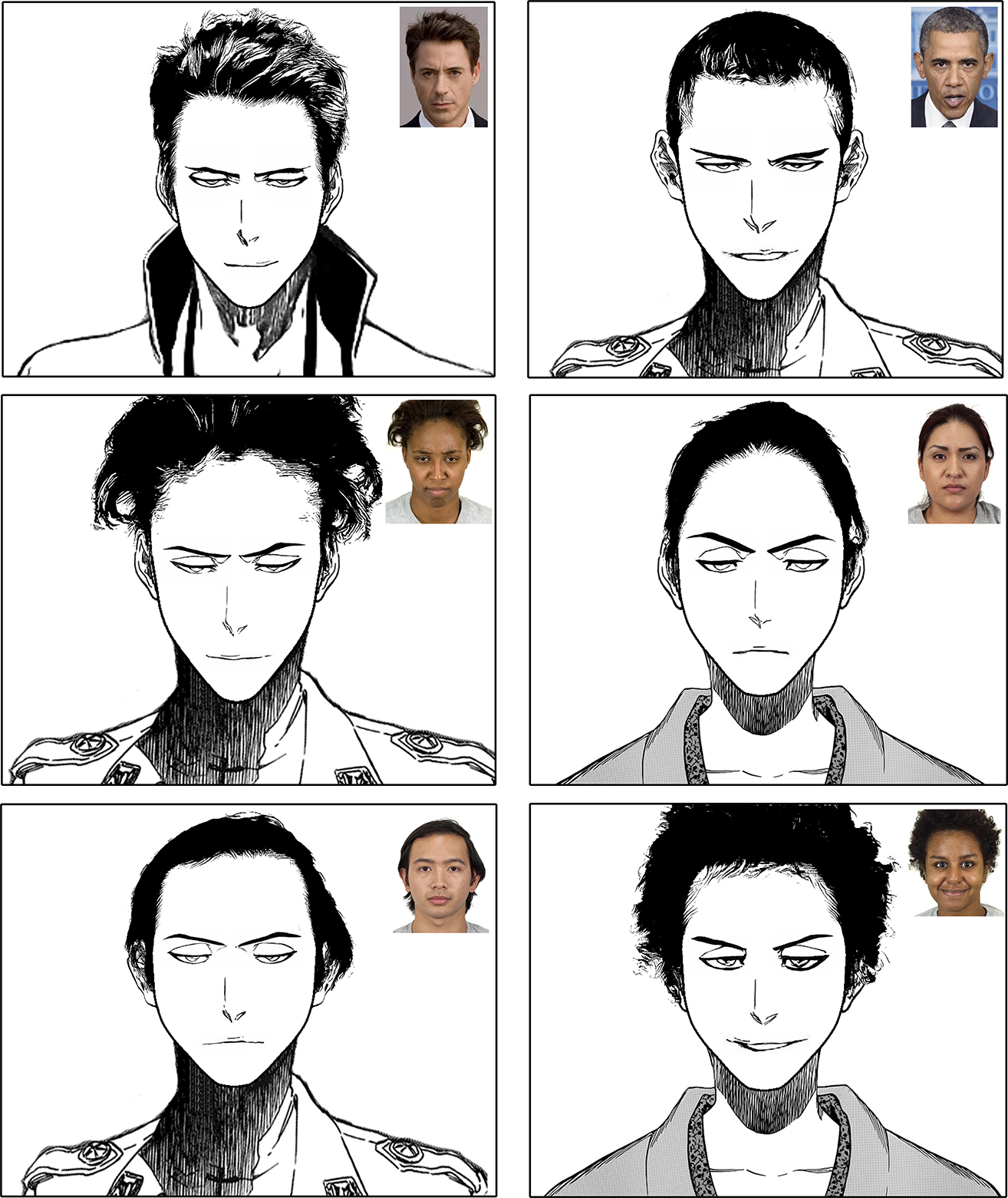}
\caption{Samples of input photos and generated manga faces}
\label{fig:demo1}
\end{figure*}
\begin{figure*}[htbp]
\centering
\includegraphics[width=6.9 in]{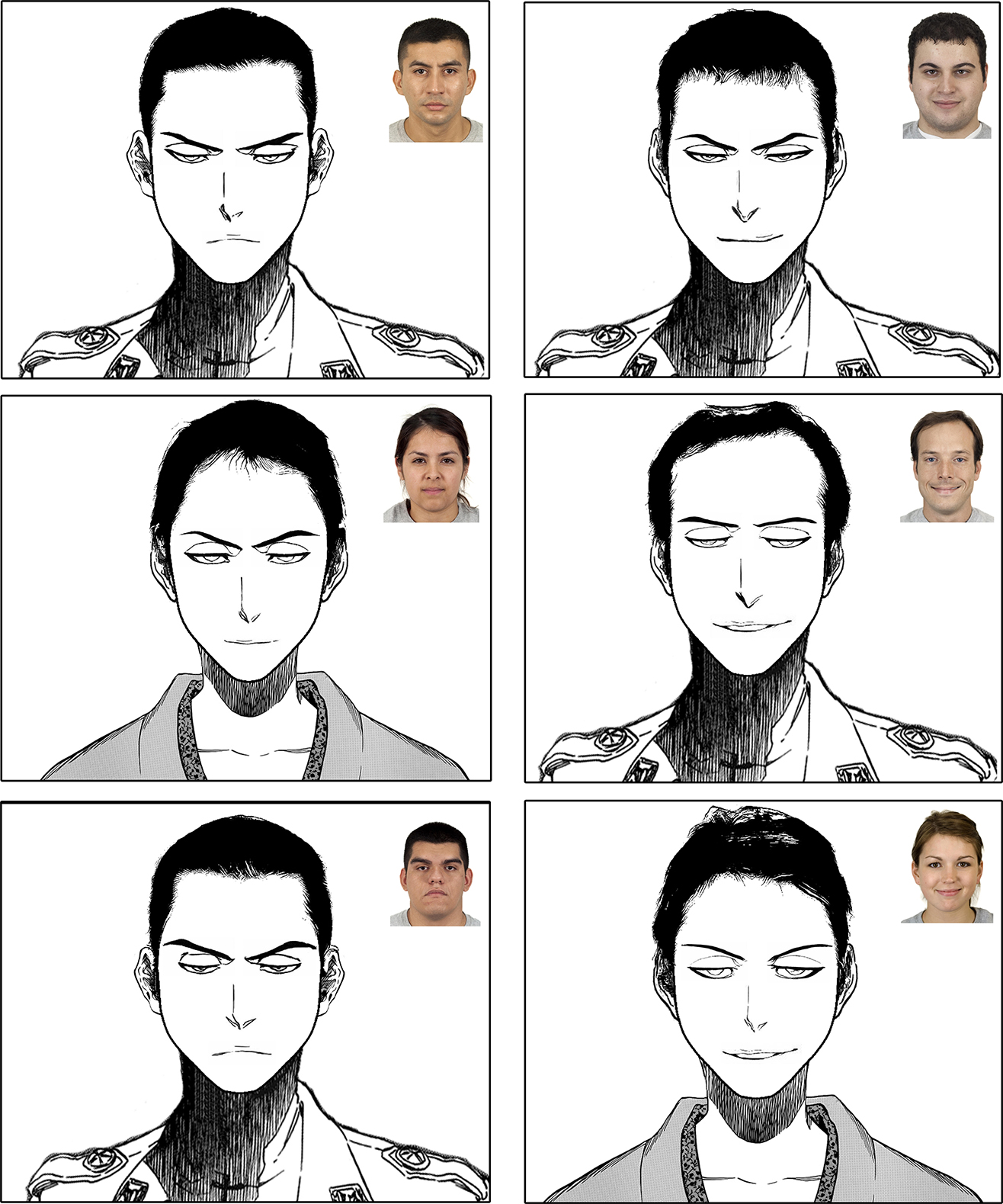}
\caption{ Samples of input photos and generated manga faces}
\label{fig:demo2}
\end{figure*}

\subsection{Generated Samples}
\label{sec:results}
  In Table \ref{tab:eye}, we show some generated samples of paired eye regions. For one people with different facial expressions, our method successfully preserves the similarities of manga eyes, and the appearances of manga eyes are adaptively changed with the facial expressions as well; for different people, our method can effectively preserve the shape of eyebrows and eyes, and further abstract them into manga style.

  Some generated samples of manga noses as shown in Figure \ref{fig:samples}. Moreover, in Figure \ref{fig:demo1} and Figure \ref{fig:demo2}, we show some manga faces with high resolution, generated for males and females.

  %We can observe that noses are insignificant to manga faces since almost all characters have a similar nose in our target manga style.

\subsection{Dataset}
\label{sec:data}
Our dataset \emph{MangaGAN-BL} is collected from a world popular manga work \emph{Bleach} \cite{bleach}. It contains manga facial features of 448 eyes, 109 noses, 179 mouths, and 106 frontal view of manga faces whose landmarks have been marked manually. Moreover, each sample of MangaGAN-BL is normalized to 256$\times$256 and optimized by cropping, angle-correction, and repairing of disturbing elements (e.g, covering of hairs, glasses, shadows).

\subsection{Failure Cases}
\label{sec:failcase}
\begin{figure}[t]
\centering
\includegraphics[width=2.8 in]{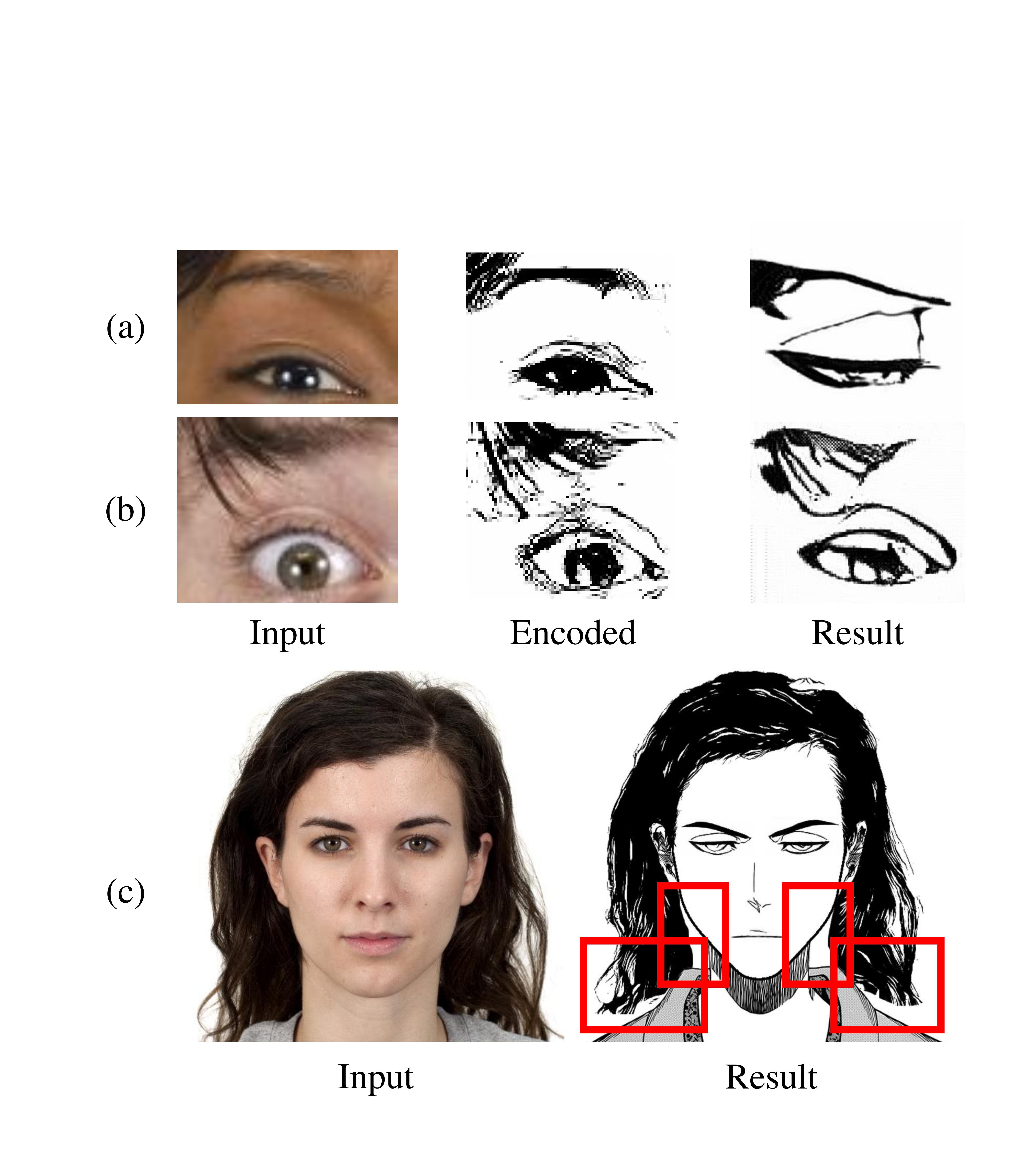}
\caption{Typical failure cases of our method. When the input eyes are close to the hair, part of the hair area may be selected into the input image, which results in some artifacts in the generated manga. }
\label{fig:failcase}
\end{figure}

Although our method can generate attractive manga faces in many cases, the network still produces some typical failure cases. As shown in Figure \ref{fig:failcase}, when the input eyes are close to the hair, part of the hair area will be selected into the input image, which results in some artifacts in the generated manga. These failure cases are caused by the incomplete content of our dataset. For example, our data for training manga eyes only include clean eye regions, thus the model cannot be adaptive to some serious interference elements (e.g., hair, glasses).

%In synthesizing manga faces, due to generated manga hairs are accurately corresponding to the input photo, in some cases, several white areas will appear (in red bounding-box) after abstracting and changing the face shape (as shown in Figure \ref{fig:failcase} (c)). Leveraging the input hair to generate a full manga hair can address this issue, however, this task requires a lot of training data (including hair's shape, texture, and color), we are collecting data for the manga hair, and plan to improve this limitation in future work.

\begin{figure}[t]
\centering
\includegraphics[width=3 in]{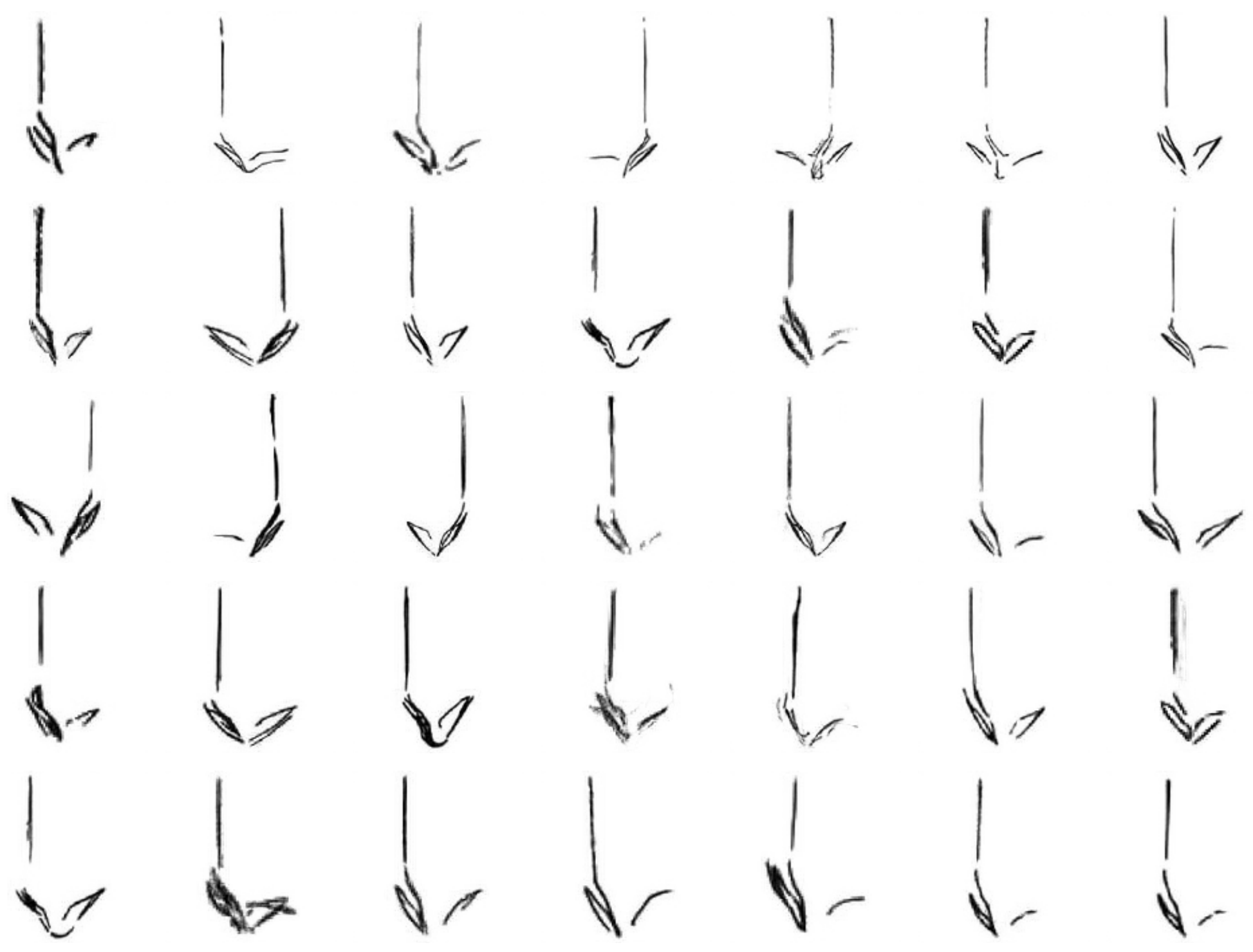}
\caption{Samples of generated manga noses.}
\label{fig:samples}
\end{figure}

{
\bibliographystyle{plain}
\bibliography{my20180302}

\end{document}